%% file: main.tex
\documentclass{article}

\usepackage{iclr2027_conference}
\iclrfinalcopy

\usepackage{times}
\usepackage{latexsym}
\usepackage[T1]{fontenc}
\usepackage[utf8]{inputenc}
\usepackage{microtype}
\usepackage{url}
\usepackage{inconsolata}
\usepackage{graphicx}
\usepackage{booktabs}
\usepackage{array}
\usepackage{tabularx}
\usepackage{multirow}
\usepackage{colortbl}
\usepackage{amsmath}
\usepackage{amssymb}
\usepackage{float}
\usepackage{placeins}
\usepackage{hyperref}
\usepackage{xcolor}
\usepackage{tcolorbox}
\tcbuselibrary{breakable}
\definecolor{preprintblue}{HTML}{365F91}
\makeatletter
\renewcommand{\section}{\@startsection{section}{1}{\z@}
  {-2.0ex plus -0.5ex minus -.2ex}{1.5ex plus .3ex minus .2ex}
  {\normalfont\large\sffamily\bfseries\color{preprintblue}\raggedright}}
\renewcommand{\subsection}{\@startsection{subsection}{2}{\z@}
  {-1.8ex plus -0.5ex minus -.2ex}{.8ex plus .2ex}
  {\normalfont\normalsize\sffamily\bfseries\color{preprintblue}\raggedright}}
\renewcommand{\subsubsection}{\@startsection{subsubsection}{3}{\z@}
  {-1.5ex plus -0.5ex minus -.2ex}{.5ex plus .2ex}
  {\normalfont\normalsize\sffamily\bfseries\color{preprintblue}\raggedright}}
\renewcommand{\paragraph}{\@startsection{paragraph}{4}{\z@}
  {1.5ex plus .5ex minus .2ex}{-1em}
  {\normalfont\normalsize\sffamily\bfseries\color{preprintblue}}}
\renewcommand{\subparagraph}{\@startsection{subparagraph}{5}{\z@}
  {1.5ex plus .5ex minus .2ex}{-1em}
  {\normalfont\normalsize\sffamily\bfseries\color{preprintblue}}}
\def\@maketitle{%
  \noindent\begin{minipage}{\textwidth}
    \includegraphics[height=0.43in]{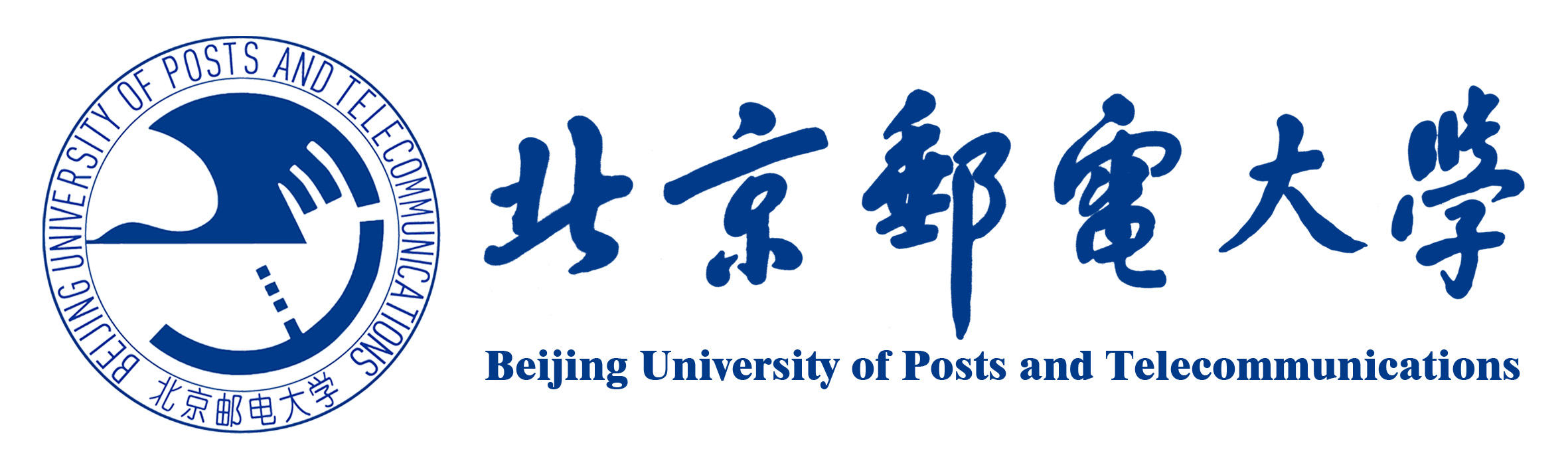}\hfill
    \includegraphics[height=0.43in]{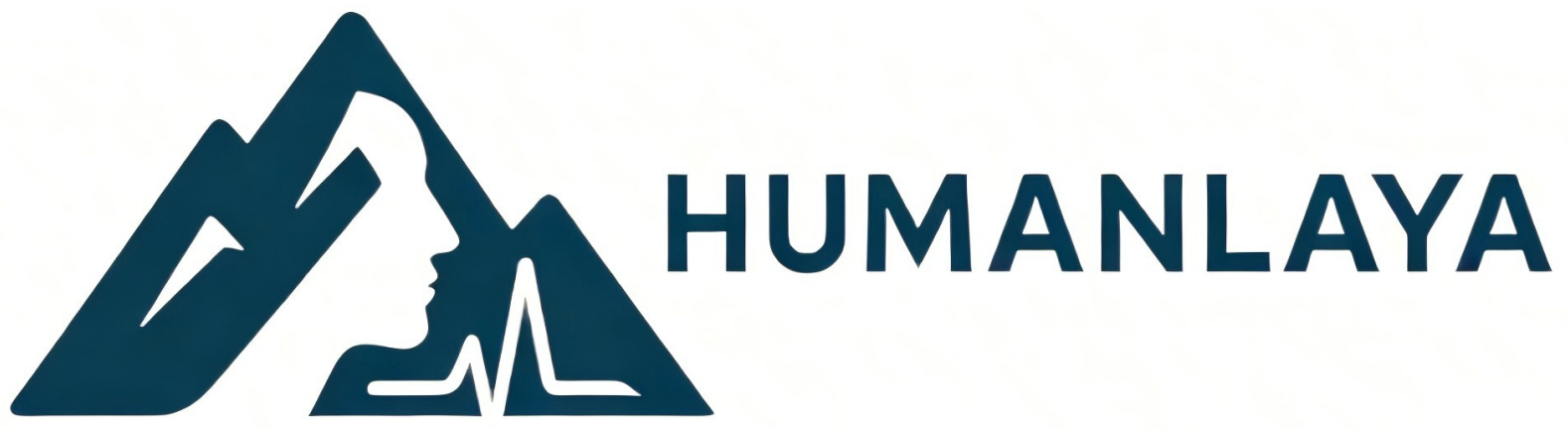}\hfill
    \includegraphics[height=0.43in,trim=38bp 170bp 403bp 130bp,clip]{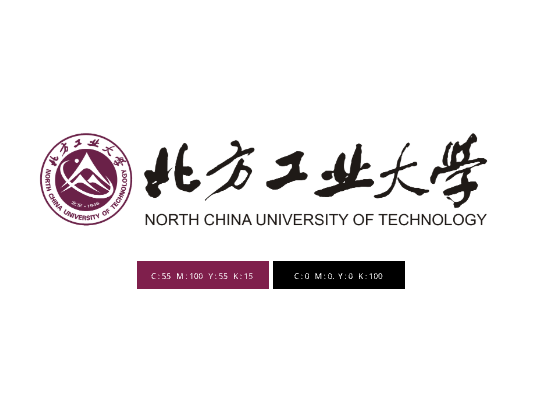}\kern3pt%
    \raisebox{0.055in}{\includegraphics[height=0.32in,trim=142bp 170bp 40bp 128bp,clip]{figures/logos/ncut_logo.png}}\hfill
    \includegraphics[height=0.43in]{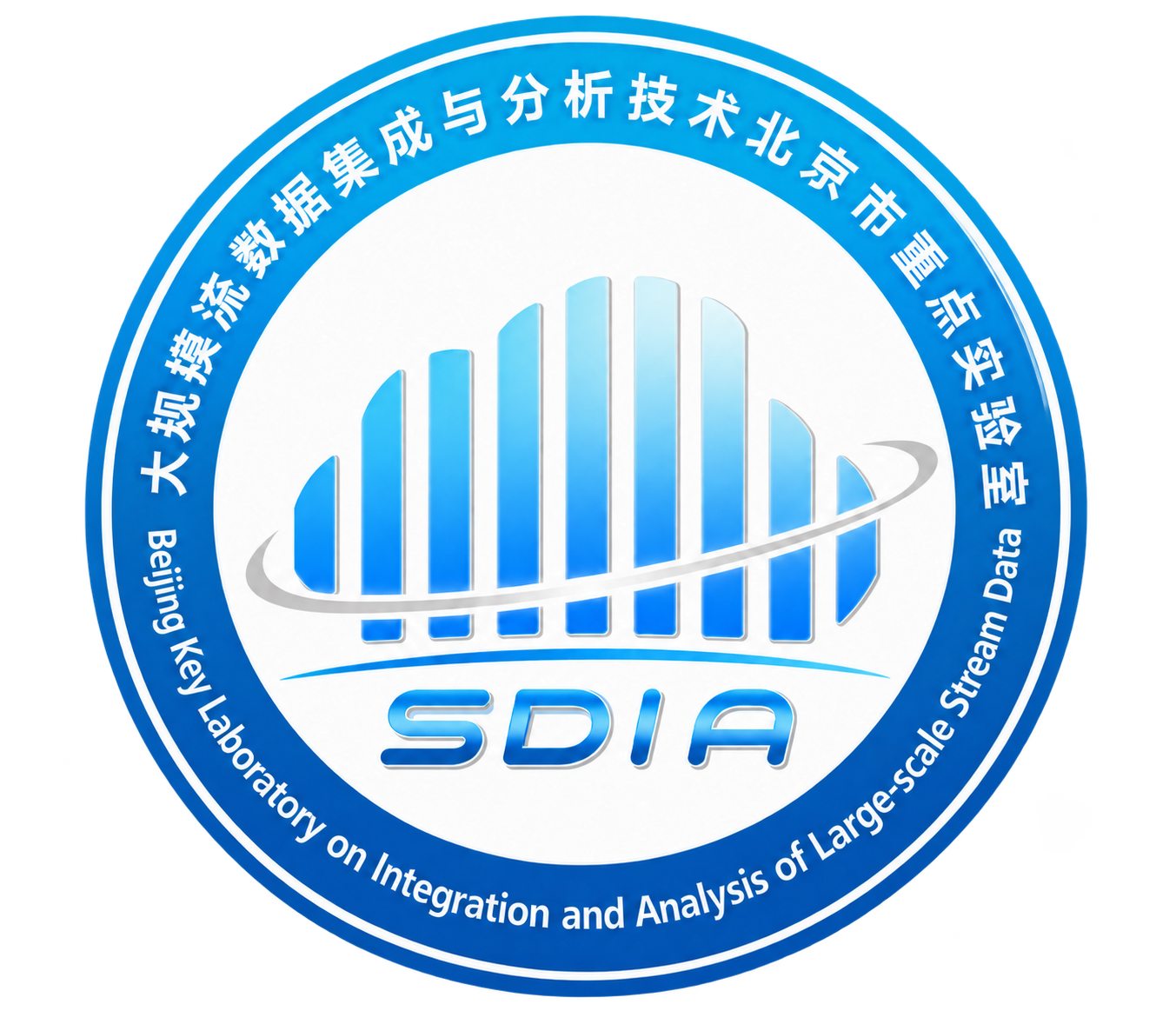}
    \par\vspace{9pt}
    {\color{preprintblue}\hrule height 0.5pt}
    \vspace{10pt}
    {\centering\fontfamily{phv}\bfseries\fontsize{16}{19}\selectfont\@title\par}
    \vspace{10pt}
    {\color{preprintblue}\hrule height 0.5pt}
    \vspace{12pt}
    {\centering\small\@author\par}
  \end{minipage}\par\vspace{12pt}}
\makeatother
\renewenvironment{abstract}{%
  \begin{tcolorbox}[colback=white,colframe=preprintblue!70,
    boxrule=0.5pt,arc=5pt,left=12pt,right=12pt,top=10pt,bottom=10pt]
  {\centering\sffamily\bfseries\large\color{preprintblue}Abstract\par}
  \vspace{6pt}\noindent\ignorespaces
}{\par\end{tcolorbox}\vspace{4pt}}
\newcounter{ragalgorithm}
\input{sections/bidirectional_contrast_values}
\input{sections/result_metric_macros}

\title{RAG-Stress: Probing the Limits of Evidence Reliance in Retrieval-Augmented Generation}

\author{
  \textbf{Shunyuan Zhou\textsuperscript{1}%
    \thanks{Equal contribution.}},
  \textbf{Hao Chen\textsuperscript{2,4}\footnotemark[1]},
  \textbf{Tianyu Wang\textsuperscript{3}},
  \\[3pt]
  \textbf{Goose Lin\textsuperscript{2}},
  \textbf{Zaiyuan Wang\textsuperscript{2}},
  \textbf{Haiying Zhao\textsuperscript{1}\thanks{Corresponding author.}}
  \\[5pt]
  \parbox{0.92\textwidth}{
    \centering
    \textsuperscript{1}Beijing University of Posts and Telecommunications
    \\[2pt]
    \textsuperscript{2}Humanlaya Data
    \quad
    \textsuperscript{3}North China University of Technology
    \\[2pt]
    \textsuperscript{4}Beijing Key Laboratory of Key Technologies for AI+ Domain Applications
  }
}

\hypersetup{hidelinks,pdfauthor={Shunyuan Zhou, Hao Chen, Tianyu Wang, Goose Lin, Zaiyuan Wang, Haiying Zhao},pdftitle={RAG-Stress: Probing the Limits of Evidence Reliance in Retrieval-Augmented Generation},pdfsubject={},pdfkeywords={}}
\begin{document}
\maketitle
% The template sets a conference-publication header in maketitle; clear it
% for arXiv without changing the official style file or claiming acceptance.
\fancyhead{}
\renewcommand{\headrulewidth}{0pt}

\begin{abstract}
\input{sections/main_abstract}
\end{abstract}

\section{Introduction}

Retrieval-augmented generation (RAG) grounds large language models in external evidence \citep{lewis2020rag,chen2025toolforgedatasynthesispipeline,su2026coalrag,anonymous2026toolforge}. Most evaluations ask whether retrieval improves answer accuracy or whether an answer is supported by the retrieved text. These criteria are insufficient when the evidence is relevant and internally coherent but supports an incompatible answer: a model can be well grounded and still be wrong. We therefore ask: \emph{when a model answers correctly without retrieval, which evidence and instruction conditions make it adopt a designated counterfactual answer?}

Two distinctions are essential: whether the model answered correctly before receiving evidence, and whether its subsequent error adopts the specific answer supported by the edit. Aggregate accuracy collapses both distinctions. RAG-Stress fixes the question and reference, edits one assertion containing the answer to support a foil of the same type, and conditions foil adoption on closed-book correctness. Responses to clean evidence provide a separate control; reference recovery on baseline failures is analyzed separately. This measures a behavioral transition, not stable parametric knowledge or, by itself, the causal effect of editing.

The intervention crosses Soft and Strict instructions with beginning, middle, and end Word Position conditions within the target text. The question is not simply whether asking models to trust documents increases compliance, but how often compliance replaces a correct baseline answer with the designated foil and how this rate varies with placement.
\input{sections/case_study}

We evaluate fifteen API models, open models, and search agents trained with RL across three QA datasets, with additional medical QA experiments in English and Chinese. Strict produces higher misleading rates than Soft across systems and datasets. Averaged equally across models, these rates follow End $>$ Beginning $>$ Middle under both instructions on all three datasets. A separate paired audit of 500 items retains 13,000 generations from two quantized 8B checkpoints (Appendix~\ref{app:paired-audit-summary}). Relative to Soft, Strict increases foil adoption on the baseline-correct subsets by 14.0 and 9.7 percentage points; intervals for reference recovery include zero. These findings indicate sensitivity to instructions and position, rather than changes to knowledge in model parameters.

Our contributions are: (1) a protocol that measures designated foil adoption conditional on baseline correctness, separating answer replacement from preexisting failure; (2) an evaluation crossing instruction and Word Position across systems, tasks, and languages; and (3) a paired audit of saved outputs separating harmful override, reference recovery, and transitions from clean to edited evidence, with uncertainty estimated over questions. The contribution is diagnostic resolution, not a new conflict phenomenon or mitigation method.

\section{Related Work}

\paragraph{Knowledge conflict and context reliance.}
Prior work studies reliance on external evidence and conflicts with prior answers \citep{mallen2023reliability,longpre2021entity,monea2024fakepedia,xie2024adaptive,farahani2024deciphering}. Entity substitutions, Fakepedia, and conflicts among sources already establish sensitivity to counterfactual evidence and source instructions \citep{he2025localsearchbenchbenchmarkingagenticsearch,zeng2026recrmbenchbenchmarkingmultidimensionalreward,he2026vistahopbenchmarkinglonghorizonvisual,he2026checkerbenchlonghorizonagentssynthesize}. Context-aware decoding changes contextual reliance \citep{shi2023trusting}; position effects in long contexts \citep{liu2024lost,he2026checkerbenchlonghorizonagentssynthesize} motivate, but do not directly test, Word Position within a document. Astute RAG, FaithfulRAG, and CARE address reliability or conflicts between facts \citep{wang2025astuterag,zhang2025faithfulrag,choi2025care}. RAG-Stress combines a crossed intervention with a fixed baseline-correct stratum and measures adoption of a designated foil. Instruction effects on conflict resolution are not themselves new.

\paragraph{RAG evaluation and robustness.}
RAG research spans retrieval, generation, and evaluation \citep{gao2023ragsurvey}; Self-RAG integrates retrieval and self-reflection \citep{asai2024selfrag}. RAGAs evaluates retrieval relevance, faithfulness, and answer quality \citep{es2024ragas}; RAGTruth annotates hallucinated claims \citep{niu2024ragtruth}. RGB and RECALL test counterfactual robustness \citep{chen2024rgb,liu2023recall}. GaRAGe, mtRAG, CRUX, and MAGIC extend evaluation to grounding, conversation, context coverage, and multi-hop conflict \citep{sorodoc2025garage,katsis2025mtrag,ju2025crux,lee2025magic}. Our complementary measurement asks whether a baseline-correct response changes to a designated foil. Neither aggregate accuracy nor grounding alone identifies this transition.

\paragraph{Adversarial evidence and memory.}
CF-RAG evaluates counterfactual queries \citep{qin2026counterfactual}, while AgentPoison, MINJA, and MemoryGraft study persistent memory or retrieval attacks \citep{chen2024agentpoison,dong2025minja,srivastava2025memorygraft}. RAG-Stress keeps the query, index, and memory fixed and measures the immediate response to one edited assertion.

\section{RAG-Stress Methodology}
\label{sec:methodology}

RAG-Stress crosses evidence condition, instruction, and Word Position (Figure~\ref{fig:pipeline}, left). Closed-book responses define model-specific strata: clean accuracy uses the full prepared set; MR uses its baseline-correct subset.

\input{sections/method_overview}

\subsection{Same-Query Evidence Intervention}
\label{sec:problem_setup}

Each example contains a question $q$, reference answer $a_q^\star$, and a clean target passage that supports the reference. We construct an edited version of that passage that supports a designated foil $a_q^-$. For open-domain QA, the foil replaces the span containing the answer with an entity of the same type; for multiple-choice QA, it is one of the provided distractor options. Within each Word Position condition, the clean/edited pair differs only in the assertion containing the answer (Appendix~\ref{app:prompts-algorithm}).

We obtain a closed-book baseline without documents. Each evidence arm crosses Soft/Strict with three Word Position conditions: the answer span occurs near the beginning, middle, or end of the target evidence text, with document order fixed. This defines twelve cells per question. Clean accuracy is measured across all prepared questions; MR is scored on baseline-correct questions. The separate paired audit retains its recorded document-slot intervention; it supports instruction contrasts, not a replication of Word Position effects.

\subsection{Baseline Strata and Evaluation Metrics}

Let $Q$ contain all prepared evaluation questions before baseline filtering. For model $M$, closed-book responses define the baseline-correct set $K_M$ and baseline-failure set $U_M$:
\begin{equation}
  K_M=\{q\in Q:\mathrm{Match}(M(q,\emptyset),a_q^\star)=1\}, \qquad U_M=Q\setminus K_M.
\end{equation}
This fixed split excludes preexisting failures from MR. It is not recomputed after an intervention. Accuracy with clean evidence uses all of $Q$:
\begin{equation}
  \mathrm{Acc}^{C}_{\pi,p}=\frac{1}{|Q|}\sum_{q\in Q}
  \mathbf{1}[\mathrm{Match}(\hat a^{C}_{q,\pi,p},a_q^\star)].
\end{equation}
Here $C$ denotes unedited evidence. Thus clean accuracy may vary with instruction $\pi$ and position $p$; it is not closed-book accuracy or accuracy on $K_M$ alone.

Let $\hat a^{E}_{q,\pi,p}$ be the response to edited evidence under instruction $\pi$ and position $p$. The main metric, misleading rate (MR), is the fraction of baseline-correct questions for which the response matches the foil but not the reference:
\begin{equation}
  \mathrm{MR}_{\pi,p}=\frac{1}{|K_M|}\sum_{q\in K_M}
  \mathbf{1}[\mathrm{Match}(\hat a_{q,\pi,p}^{E},a_q^-)
  \land \neg\mathrm{Match}(\hat a_{q,\pi,p}^{E},a_q^\star)].
\end{equation}
All rates are reported as percentages; double matches do not count as MR. MR is not the clean-to-edited accuracy drop: $K_M$ need not be clean-evidence-correct, and foil adoption is only one error type. Direct transitions require paired clean/edited outputs (Appendix~\ref{sec:direct-transitions}). Audit HOR averages MR over positions; BCR measures reference recovery with clean evidence on $U_M$. These conditional rates remain separate from accuracy on the full set and from each other.

\subsection{Paired Contrasts and Uncertainty}
Instruction contrasts subtract Soft MR from Strict MR after averaging positions within question. Position contrasts pair the same questions and instruction. The supplementary audit uses 20,000 bootstrap resamples of questions for pointwise 95\% percentile intervals, retaining all repeated cells together. These intervals are not multiplicity-adjusted; primary reporting does not imply preregistration. The QA position check uses 2,000 resamples. Neither set of intervals applies to the rate tables for fifteen systems (Algorithm~\ref{alg:rag-stress}; Appendices~\ref{sec:bidirectional-contrasts} and~\ref{app:robustness-protocols}).

\section{Experiments}
\label{sec:experiments}

\subsection{Datasets and Experimental Setup}

We evaluate TriviaQA-RC \citep{joshi2017triviaqa}, HotpotQA \citep{yang-etal-2018-hotpotqa}, and SearchQA \citep{dunn2017searchqa}. The model set contains seven API systems, four open-source instruction models \citep{grattafiori2024llama3,jiang2023mistral,yang2025qwen3}, and four RL-trained search agents \citep{jin2025searchr1trainingllmsreason,sun2026zerosearchincentivizesearchcapability,chu2026redsearcherscalablecostefficientframework,xia2026searchp1pathcentricrewardshaping}. Live search is disabled; evidence is fixed rather than retrieved by each system. Clean responses are evaluated on the full prepared set under all six instruction--position conditions. Closed-book filtering defines each model's eligible subset for edited-evidence MR, not the denominator of clean accuracy.

Closed-book prompting requests a short answer without explanation. Strict prompting makes the documents the primary source of truth even when they appear mistaken; Soft prompting permits prior knowledge to override conflicting documents. Thus the prompt contrast intentionally changes the conflict-resolution policy rather than paraphrasing the same instruction (Appendix~\ref{sec:prompt-templates}).

\subsection{Baselines and Evaluation}

The primary outcome is model-conditional MR, with clean accuracy, conflict follow, gold retention, and hedge rate as diagnostics. Responses are normalized by case, punctuation, articles, and answer prefixes; gold/foil double matches do not count as MR. Main runs use greedy decoding and a short-answer generation cap. The contrasts in the rate tables for fifteen systems are descriptive. Paired intervals are reported separately for the 300-question QA checks and the 500-item audit. Cross-dataset and model-family comparisons remain descriptive because their retained question sets differ.

\subsection{Results and Behavioral Interpretation}
\label{sec:results}

Table~\ref{tab:rs-triviaqa} gives the main TriviaQA-RC grid; Tables~\ref{tab:rs-hotpotqa} and~\ref{tab:rs-searchqa} in Appendix~\ref{app:full-experimental-results} give HotpotQA and SearchQA. Quantitative summaries below use these tables. Strict/Soft cells pair bold clean accuracy on the full set with teal $\downarrow$ MR on each model's baseline-correct subset. Each $\Delta$ cell pairs the corresponding clean-accuracy and MR contrasts (pp); the arrow denotes MR, not accuracy loss. Unless qualified, textual contrasts refer to MR. Group means weight models equally and are rounded last (Appendix~\ref{sec:aggregation-details}).

\input{sections/rs_table_triviaqa}

\paragraph{Instructions act as a policy for selecting sources.}
Strict MR exceeds Soft MR in every reported cell of the main tables. Within each comparison, question, evidence, and placement are fixed; the policy for source priority changes. The gap is therefore consistent with evidence following that depends on the policy, rather than different retrieved content. Its direction is expected from the instructions; the diagnostic result is the frequency of designated-foil adoption among baseline-correct answers. Output differences alone do not identify internal attention, the relative weight of memory and context, or erasure of a stored fact.

\paragraph{Word Position changes evidence use.}
Averaging equally across the fifteen models gives End $>$ Beginning $>$ Middle under both instructions on all three datasets. Strict MR at these positions is 37.0/31.4/30.3\% on TriviaQA-RC, 41.3/35.8/32.8\% on HotpotQA, and 35.0/29.3/26.9\% on SearchQA. This is an aggregate pattern, not a universal per-model ordering. Later placement may change the answer span's salience, but does not identify an attention mechanism or document-order effect. Section~\ref{sec:word-position-validation} provides a paired position check with uncertainty estimates.

\paragraph{Dataset structure may shape override.}
HotpotQA has higher MR averaged across models than TriviaQA-RC and SearchQA under both instructions. One explanation is that a false intermediate assertion can support an internally coherent but incorrect reasoning chain, not just replace its final answer. The present comparison does not isolate that mechanism: evidence style, difficulty, and baseline-correct cohorts also differ. Testing it would require matched bridge-versus-terminal edits, rather than treating a cross-dataset difference as a causal multi-hop effect.

\paragraph{Model families differ in default evidence reliance.}
API systems have lower group mean MR than open 7--9B models; RL-trained search systems have the highest Soft MR and smallest mean instruction gap (SearchQA: 10.1 vs.\ 10.3 pp for APIs). This is consistent with greater evidence reliance under Soft, not an isolated RL-training effect. Each model's MR conditions on a different $K_M$: low MR need not imply higher accuracy on the full set. Cross-model rankings therefore require both baseline coverage and matched-cohort controls; a small instruction gap alone is not robustness.

\paragraph{Clean controls separate override from general degradation.}
Table~\ref{tab:rs-triviaqa} displays clean accuracy on the full set alongside baseline-conditioned edited MR; these have different denominators and are not subtracted. The separate paired audit compares clean and edited responses on matched questions to assess reference retention. Appendices~\ref{sec:clean-edited-control} and~\ref{sec:direct-transitions} report matched foil adoption and direct clean-gold to edited-foil transitions; Appendix~\ref{sec:gold-score-interaction} compares gold retention on a common cohort.

\subsection{Ablations and Robustness Checks}
\label{sec:ablation}

We check paired position differences, sensitivity to prompt wording, and whether the instruction effect extends across tasks and languages. Further checks address edit validity, matcher errors, and decoding profiles.

\input{sections/word_position_validation}

\input{sections/prompt_paraphrase_robustness}

\subsubsection{Cross-Lingual Generalization on Medical QA}
\label{sec:crosslingual}

To assess whether the instruction contrast extends to medical QA in two languages, we use English USMLE and simplified-Chinese MCMLE questions from MedQA \citep{jin2020diseasedoespatienthave}, with MedRAG evidence \citep{xiong2024benchmarkingretrievalaugmentedgenerationmedicine}. Each language contains 1,000 sampled questions. Foils are provided distractor options; MR uses each model's subset answered correctly without retrieval ($n_K$ in Table~\ref{tab:crosslingual}). Group means weight models equally. The two languages contain different questions, not translations of a paired test set.

\input{sections/rs_table_multilingual}

\paragraph{The instruction effect is consistent across languages.}
Strict MR exceeds Soft MR for all 15 systems in both languages. Group-averaged $\Delta_{\mathrm{instr}}$ is similar between English and Chinese (API-based: 12.8 vs.\ 12.4; open-source: 16.9 vs.\ 18.3; RL-based: 9.8 vs.\ 9.3). The instruction effect is therefore consistent across the two language settings.

\paragraph{Language differences covary with closed-book coverage.}
The language effect $\Delta_{\mathrm{lang}}$ covaries with closed-book coverage. English-centric open models retain fewer Chinese questions and show higher Chinese MR, whereas Qwen3-8B and several Chinese-developed API models show the opposite direction. Because USMLE and MCMLE contain different questions, this pattern is descriptive and cannot isolate language proficiency from item composition.

\paragraph{High Soft MR in search agents.}
RL-based systems have the highest Soft MR (40.1\% English and 41.8\% Chinese on average) and the smallest group-averaged $\Delta_{\mathrm{instr}}$. This indicates substantial foil following even under Soft prompting, but comparisons across different systems do not isolate the effect of RL training.

\subsubsection{Counterfactual Evidence and Matcher Validation}
\label{sec:ablation-validation}

\paragraph{Human validation of counterfactual evidence.}
We sample 200 edits from each of TriviaQA-RC, HotpotQA, SearchQA, and the two MedQA languages (1,000 total). Six annotators cross-validate the edits: two independently label each edit for foil support, relation preservation, fluency, and leakage of the reference answer, with disagreements adjudicated by a third.

\input{sections/ablation_evidence_validation_table}

In Table~\ref{tab:evidence-validation}, 88.1\% of edits pass all criteria ($\kappa=0.77$). Relation preservation is the main bottleneck for HotpotQA and Chinese MedQA; SearchQA has the lowest fluency. Filtering increases mean $\Delta_{\mathrm{instr}}$ from 12.3 to 12.8 pp within this comparison, preserving its direction. These are not model group means in the main tables, and filtering changes item composition.

\paragraph{Matcher error analysis.}
We sample 450 outputs stratified by matcher label (Gold, Foil-only, and Unmatched; 150 each) across datasets, models, instructions, and positions. Human labels identify the committed answer. For matcher--human agreement, human Hedge and Other labels map to Unmatched; the displayed counts yield Cohen's $\kappa=0.91$, not an inter-annotator coefficient.

\input{sections/ablation_matcher_table}

Table~\ref{tab:matcher-analysis} gives 94.2\% agreement and 95.3\% Foil-only precision on the sample balanced by label. Its 26 disagreements mainly involve aliases, formatting, hedging, or multiple entities. Sampling stratified by label prevents interpreting these statistics as agreement across the population. This diagnoses scoring errors, not benchmark MR sensitivity to an alternative judge.

\subsubsection{Decoding Profiles}
\label{sec:ablation-decoding}

Table~\ref{tab:decoding-stability} reports six systems on TriviaQA-RC under temperature sampling (temperature 0.7, top-$p$ 0.9, five seeds) and greedy decoding with a 256-token output cap. These are separate decoding profiles, not a matched single-factor contrast against the runs in the main tables.

\input{sections/ablation_decoding_table}

Both profiles retain positive mean instruction contrasts for all six systems. Reported seed standard deviations are at most 1.6 pp, and the Strict-MR ordering is identical between the sampling means and long-generation values (Kendall $\tau=1.00$). This is a descriptive consistency check: it does not isolate a sampling or output-length effect, nor measure changes relative to Table~\ref{tab:rs-triviaqa}.

\paragraph{Ablation summary.}
The checks address different threats: task transfer, edit validity, scoring error, decoding, and wording. Reported instruction contrasts remain positive; the paired position check supports higher End MR under Strict, not the full aggregate ordering. These are complementary diagnostics, not interchangeable replications or isolated tests of every causal factor.

\FloatBarrier

\section{Conclusion}

RAG-Stress measures a specific reliability failure: adoption of an evidence-supported foil after a correct closed-book response. Across fifteen systems and three QA datasets, the reported mean Strict--Soft MR gap is 10.9--13.5 pp; rates averaged across models follow End $>$ Beginning $>$ Middle under both instructions. Medical QA and decoding checks extend the instruction pattern beyond one setting. A separate audit of saved outputs supports increased harmful override without establishing a gain in beneficial correction. These findings distinguish compliance with a policy for source priority from factual reliability; they do not identify an internal attention mechanism or a superior mitigation. The practical evaluation lesson is to report baseline coverage, foil adoption, and matched clean/edited transitions separately, rather than infer robustness from aggregate accuracy or document adherence alone.

\FloatBarrier
\label{page:main-end}

\bibliographystyle{iclr2027_conference}
\bibliography{custom}

\appendix
\raggedbottom
\tcbset{breakable}
\input{sections/prompt_algorithm_appendix}

\section{Supplementary Paired Audit}
\label{app:paired-audit-summary}

The supplementary experiment uses the same 500 prepared TriviaQA-RC items for Llama-3-8B-Instruct \citep{grattafiori2024llama3} and Qwen3-8B \citep{yang2025qwen3}. Selection precedes inference: 788 candidates from the historically Llama-selected 1,000-item pool yield 500 lexically eligible pairs, without replacement. Each model receives one closed-book request and twelve clean/edited grid requests per item, totaling 6,500 calls per model. The baseline strata defined for each model are computed afterward, not used to skip calls. Both checkpoints use local MLX inference, four-bit affine quantization, greedy decoding, a 20-token output cap, and a 1,536-token input ceiling. Unlike the main Word Position intervention, this audit moves the entire target passage to document slot 1, 2, or 4. Its position contrasts describe document order, not word placement within the text. Scoring uses normalized exact matching. Appendix~\ref{app:bidirectional-record} records tokenizer/checkpoint fingerprints, exclusions, realized token lengths, and execution limitations. No parameter updates, index rebuilding, or reranking are performed. The main evaluation and this audit are separate cohorts: rates in the main tables and human-validation summaries do not certify the audit items. The audit verifies lexical construction and response scoring rather than independent semantic validity; its human-review fields remain unfilled. Rates are interpreted as output behavior.

\input{sections/bidirectional_results}

\FloatBarrier
\input{sections/paired_audit_analysis}
\FloatBarrier
\input{sections/bidirectional_execution}
\input{sections/full_experimental_results}

\FloatBarrier
\input{sections/robustness_protocol_appendix}
\end{document}

%% file: sections/bidirectional_contrast_values.tex
\newcommand{\LlamaHORPromptContrast}{$14.0\ [11.1,17.0]$}
\newcommand{\LlamaBCRPromptContrast}{$-2.1\ [-5.1,1.0]$}
\newcommand{\LlamaHORPositionContrast}{$3.8\ [-0.3,8.0]$}
\newcommand{\LlamaHORSoftPositionContrast}{$6.2\ [1.8,10.7]$}
\newcommand{\LlamaHORPositionInteraction}{$-2.4\ [-7.4,2.7]$}
\newcommand{\LlamaGoldCleanPromptContrast}{$-2.4\ [-4.1,-0.7]$}
\newcommand{\LlamaGoldEditedPromptContrast}{$-11.8\ [-14.1,-9.6]$}
\newcommand{\LlamaGoldScoreInteraction}{$-9.4\ [-11.9,-6.9]$}
\newcommand{\LlamaDirectGoldFoilTransition}{1123/2028 (55.4\%)}
\newcommand{\LlamaGoldOriginFoilTransition}{64.5\%}
\newcommand{\LlamaStrictCleanRow}{0.0\% & 64.5\% & $64.5\ [59.9,69.1]$}
\newcommand{\LlamaSoftCleanRow}{0.3\% & 50.5\% & $50.2\ [45.5,55.0]$}
\newcommand{\QwenHORPromptContrast}{$9.7\ [6.8,12.6]$}
\newcommand{\QwenBCRPromptContrast}{$0.1\ [-1.5,1.8]$}
\newcommand{\QwenHORPositionContrast}{$11.6\ [7.7,15.8]$}
\newcommand{\QwenHORSoftPositionContrast}{$8.1\ [3.9,12.7]$}
\newcommand{\QwenHORPositionInteraction}{$3.5\ [-1.9,8.9]$}
\newcommand{\QwenGoldCleanPromptContrast}{$-0.4\ [-1.7,0.9]$}
\newcommand{\QwenGoldEditedPromptContrast}{$-5.7\ [-7.5,-4.1]$}
\newcommand{\QwenGoldScoreInteraction}{$-5.3\ [-7.5,-3.3]$}
\newcommand{\QwenDirectGoldFoilTransition}{872/1554 (56.1\%)}
\newcommand{\QwenGoldOriginFoilTransition}{63.5\%}
\newcommand{\QwenStrictCleanRow}{0.0\% & 63.8\% & $63.8\ [58.3,69.2]$}
\newcommand{\QwenSoftCleanRow}{0.0\% & 54.2\% & $54.2\ [48.5,59.8]$}

%% file: sections/result_metric_macros.tex
\definecolor{resultdown}{HTML}{008D9C}
\newcommand{\cleanmr}[2]{%
  \mbox{\textbf{#1}\raisebox{-0.35ex}{\fontsize{5}{5}\selectfont
    \textcolor{resultdown}{\ensuremath{\downarrow}#2}}}}
\newcommand{\cleandelta}[2]{%
  \mbox{\textbf{#1}\raisebox{-0.35ex}{\fontsize{5}{5}\selectfont
    \textcolor{resultdown}{\ensuremath{\downarrow}#2}}}}

%% file: sections/main_abstract.tex
Following retrieved evidence does not guarantee factual correctness: misleading evidence can induce a model to replace an answer it previously gave correctly. Standard accuracy measures obscure this behavior by combining answer replacement with preexisting errors. We introduce RAG-Stress, a controlled diagnostic protocol for examining the limits of evidence reliance in retrieval-augmented generation. The protocol holds the question and reference answer fixed, edits one assertion to support a designated incorrect answer, and crosses two source priority policies with three positions of the answer span within the evidence text. We measure misleading rate (MR) on each model's subset of questions answered correctly without retrieval, alongside clean accuracy on the full evaluation set. We evaluate fifteen systems spanning API models, open models, and search agents trained with reinforcement learning on TriviaQA-RC, HotpotQA, and SearchQA, with additional English and Chinese MedQA evaluations. Instructions that prioritize documents consistently produce higher MR than those permitting reliance on prior knowledge. Averaged over models and positions, the gap ranges from 10.9 to 13.5 percentage points across the three QA datasets. Mean MR follows End $>$ Beginning $>$ Middle under both policies, although individual models do not uniformly follow this ordering. A separate paired audit of 500 questions and two checkpoints supports increased harmful override without establishing a corresponding improvement in beneficial correction. These findings distinguish evidence adherence from factual reliability and motivate evaluating whether retrieved evidence preserves, replaces, or corrects a model's answers.

%% file: sections/case_study.tex
\begin{figure}[H]
\centering
\includegraphics[width=\linewidth]{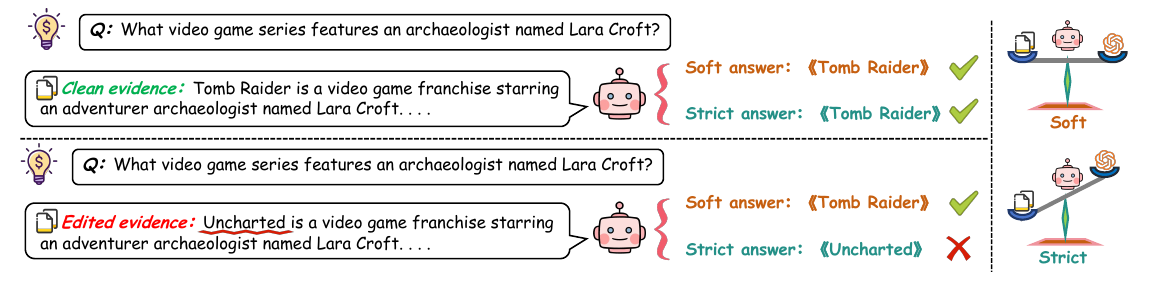}
\caption{One saved trajectory: the same edit, two instructions, different answers. With clean evidence, both soft and strict prompting return the reference (Tomb Raider); with the edited evidence, soft prompting keeps the reference while strict prompting adopts the foil (Uncharted). The closed-book answer is also Tomb Raider.}
\label{fig:case-study}
\end{figure}

%% file: sections/method_overview.tex
\begin{figure}[H]
\centering
\definecolor{methodblue}{HTML}{244962}
\definecolor{methodgreen}{HTML}{247B75}
\definecolor{methodred}{HTML}{BB5847}
\definecolor{methodink}{HTML}{243642}
\includegraphics[width=\textwidth]{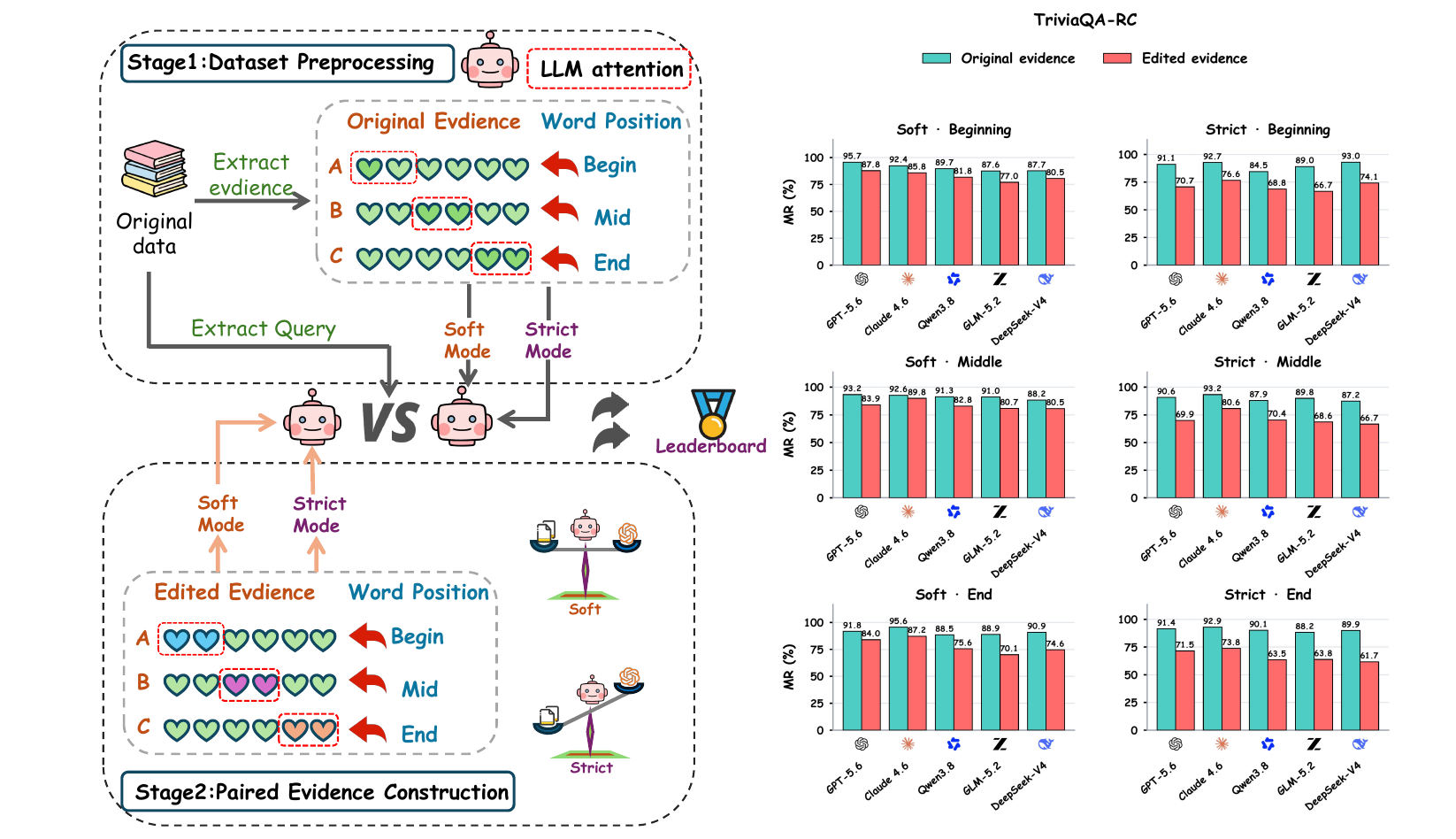}
\caption{RAG-Stress overview. \textbf{Left}: paired evidence construction and the instruction--Word Position design. \textbf{Right}: original- and edited-evidence results for five API models on TriviaQA-RC, organized into six instruction--position panels.}
\label{fig:pipeline}
\end{figure}

%% file: sections/rs_table_triviaqa.tex
% Main results table (layout adapted from CheckerBench Table 2).
% Self-contained: logo macros below; icons in figures/model-icons and figures/harness-icons.
% Requires: graphicx, xcolor, colortbl, booktabs, tabularx, array.
\providecommand{\brandlabel}[2]{%
  \raisebox{-0.15em}{\includegraphics[height=1em,width=1em,keepaspectratio]{#1}}%
  \hspace{0.35em}#2}
\providecommand{\modelname}[2]{\brandlabel{figures/model-icons/model-#1.png}{#2}}
\providecommand{\harnessname}[2]{\brandlabel{figures/harness-icons/#1.png}{#2}}

% Inline changes follow the supplied reference: orange increases, teal decreases.
\begin{table*}[tp]
\centering
\scriptsize
\definecolor{resultheader}{HTML}{DDE6EE}
\definecolor{resultstripe}{HTML}{F5F7F9}
\definecolor{resultgroup}{HTML}{EBF0F4}
\definecolor{resultup}{HTML}{D95638}
\definecolor{resultdown}{HTML}{008D9C}
\definecolor{datasetname}{HTML}{365F91}
\newcommand{\metricheader}[1]{%
  \begin{tabular}[c]{@{}c@{}}#1\end{tabular}}
% Reserve the same space for every value and subscript so decimals align.
\newcommand{\metricvalue}[2]{%
  \mbox{#1#2}}
% Column header centred over the value (not the value+subscript block).
\newcommand{\colhead}[1]{%
  \mbox{#1}}
\newcommand{\metricchange}[3]{%
  \raisebox{-0.35ex}{\fontsize{5}{5}\selectfont
    \textcolor{#1}{\ensuremath{#2}#3}}}
\newcommand{\deltaup}[1]{\metricchange{resultup}{\uparrow}{#1}}
\newcommand{\deltadown}[1]{\metricchange{resultdown}{\downarrow}{#1}}
\newcommand{\deltaequal}{\metricchange{gray}{=}{0.00}}
\caption{TriviaQA-RC. Beginning/Middle/End: Word Position within the target evidence text. Strict/Soft cells: bold full-set clean accuracy; $\downarrow$ MR on closed-book-correct items (\%). Each $\Delta$ cell: bold clean-accuracy contrast and $\downarrow$ MR contrast, both Strict minus Soft (pp); Mean averages positions. The arrow identifies the MR component, not an accuracy drop or a sign.}
\label{tab:rs-triviaqa}
\setlength{\tabcolsep}{2pt}
\renewcommand{\arraystretch}{1.22}
\begin{tabularx}{\textwidth}{l*{10}{>{\centering\arraybackslash}X}}
\toprule[1.5pt]
\rowcolor{resultheader}
\brandlabel{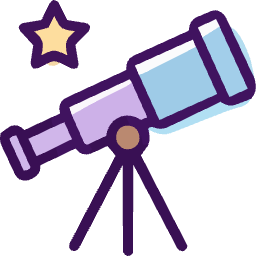}{\textcolor{datasetname}{\textbf{\textit{TriviaQA-RC}}}} & \multicolumn{3}{c}{\textbf{Beginning}} & \multicolumn{3}{c}{\textbf{Middle}} & \multicolumn{3}{c}{\textbf{End}} & \colhead{\textbf{Mean}} \\
\arrayrulecolor{gray!60}\cmidrule(lr){2-4}\cmidrule(lr){5-7}\cmidrule(lr){8-10}\cmidrule(lr){11-11}\arrayrulecolor{black}
\rowcolor{resultheader}
\textbf{Model} & \colhead{\textbf{Strict}} & \colhead{\textbf{Soft}} & \colhead{$\boldsymbol{\Delta}_{\mathbf{instr}}$} & \colhead{\textbf{Strict}} & \colhead{\textbf{Soft}} & \colhead{$\boldsymbol{\Delta}_{\mathbf{instr}}$} & \colhead{\textbf{Strict}} & \colhead{\textbf{Soft}} & \colhead{$\boldsymbol{\Delta}_{\mathbf{instr}}$} & \colhead{$\overline{\boldsymbol{\Delta}}_{\mathbf{instr}}$} \\
\midrule[0.6pt]
\rowcolor{resultgroup}
\multicolumn{11}{c}{\textbf{\textit{API-based}}} \\
\modelname{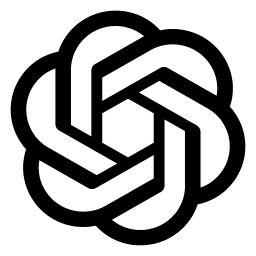}{GPT-5.6} & \cleanmr{91.1}{20.4} & \cleanmr{95.7}{7.9} & \cleandelta{-4.6}{12.5} & \cleanmr{90.6}{20.7} & \cleanmr{93.2}{9.3} & \cleandelta{-2.6}{11.4} & \cleanmr{91.4}{19.9} & \cleanmr{91.8}{7.8} & \cleandelta{-0.4}{12.1} & \cleandelta{-2.5}{12.0} \\
\rowcolor{resultstripe}
\modelname{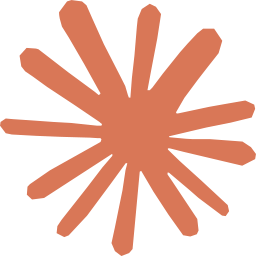}{Claude Opus 4.6} & \cleanmr{92.7}{16.1} & \cleanmr{92.4}{6.6} & \cleandelta{0.3}{9.5} & \cleanmr{93.2}{12.6} & \cleanmr{92.6}{2.8} & \cleandelta{0.6}{9.8} & \cleanmr{92.9}{19.1} & \cleanmr{95.6}{8.4} & \cleandelta{-2.7}{10.7} & \cleandelta{-0.6}{10.0} \\
\modelname{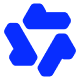}{Qwen3.8-Max} & \cleanmr{84.5}{15.7} & \cleanmr{89.7}{7.9} & \cleandelta{-5.2}{7.8} & \cleanmr{87.9}{17.5} & \cleanmr{91.3}{8.5} & \cleandelta{-3.4}{9.0} & \cleanmr{90.1}{26.6} & \cleanmr{88.5}{12.9} & \cleandelta{1.6}{13.7} & \cleandelta{-2.3}{10.2} \\
\rowcolor{resultstripe}
\modelname{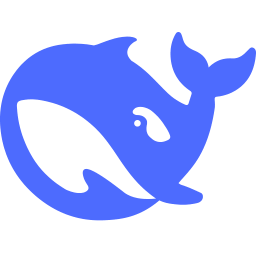}{DeepSeek-V4-Pro} & \cleanmr{93.0}{18.9} & \cleanmr{87.7}{7.2} & \cleandelta{5.3}{11.7} & \cleanmr{87.2}{20.5} & \cleanmr{88.2}{7.7} & \cleandelta{-1.0}{12.8} & \cleanmr{89.9}{28.2} & \cleanmr{90.9}{16.3} & \cleandelta{-1.0}{11.9} & \cleandelta{1.1}{12.1} \\
\modelname{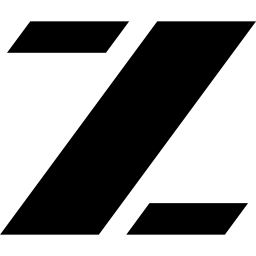}{GLM-5.2} & \cleanmr{89.0}{22.3} & \cleanmr{87.6}{10.6} & \cleandelta{1.4}{11.7} & \cleanmr{89.8}{21.2} & \cleanmr{91.0}{10.3} & \cleandelta{-1.2}{10.9} & \cleanmr{88.2}{24.4} & \cleanmr{88.9}{18.8} & \cleandelta{-0.7}{5.6} & \cleandelta{-0.2}{9.4} \\
\rowcolor{resultstripe}
\modelname{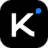}{Kimi-K3} & \cleanmr{91.0}{18.9} & \cleanmr{87.9}{5.9} & \cleandelta{3.1}{13.0} & \cleanmr{89.3}{22.0} & \cleanmr{89.3}{7.1} & \cleandelta{0.0}{14.9} & \cleanmr{98.0}{26.1} & \cleanmr{91.0}{9.5} & \cleandelta{7.0}{16.6} & \cleandelta{3.4}{14.8} \\
\modelname{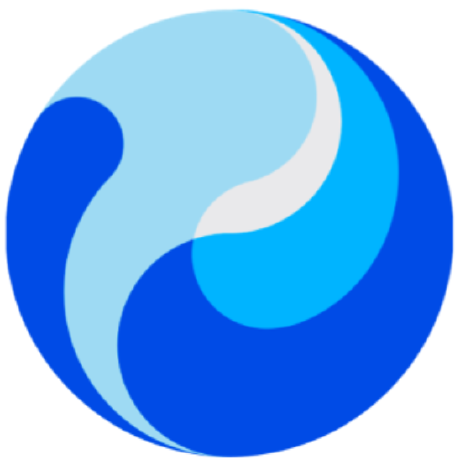}{HY4-Preview} & \cleanmr{89.2}{30.1} & \cleanmr{82.1}{14.1} & \cleandelta{7.1}{16.0} & \cleanmr{85.7}{26.8} & \cleanmr{89.5}{10.3} & \cleandelta{-3.8}{16.5} & \cleanmr{89.7}{32.5} & \cleanmr{91.3}{14.9} & \cleandelta{-1.6}{17.6} & \cleandelta{0.6}{16.7} \\
\rowcolor{resultgroup}
\textbf{Average} & \cleanmr{90.1}{20.3} & \cleanmr{89.0}{8.6} & \cleandelta{1.1}{11.7} & \cleanmr{89.1}{20.2} & \cleanmr{90.7}{8.0} & \cleandelta{-1.6}{12.2} & \cleanmr{91.5}{25.3} & \cleanmr{91.1}{12.7} & \cleandelta{0.3}{12.6} & \cleandelta{-0.1}{12.2} \\
\midrule[0.4pt]
\rowcolor{resultgroup}
\multicolumn{11}{c}{\textbf{\textit{Open-source}}} \\
\modelname{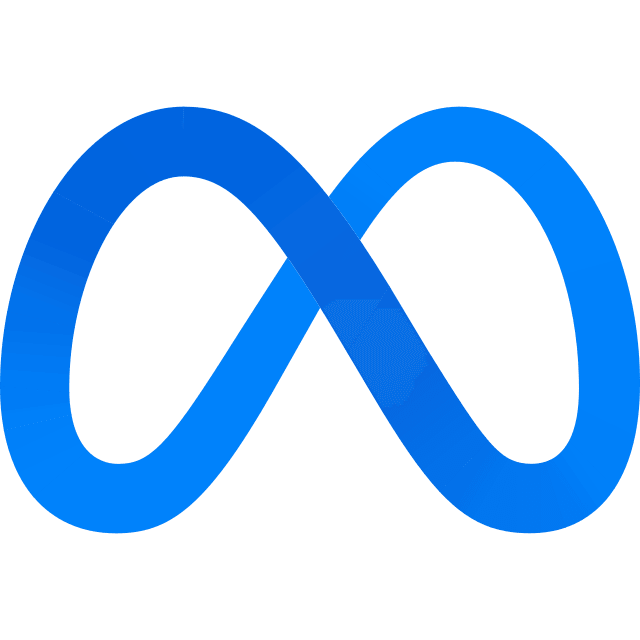}{Llama-3-8B} & \cleanmr{75.9}{35.0} & \cleanmr{78.2}{23.2} & \cleandelta{-2.3}{11.8} & \cleanmr{79.4}{35.4} & \cleanmr{76.3}{24.3} & \cleandelta{3.1}{11.1} & \cleanmr{78.7}{44.5} & \cleanmr{76.0}{30.6} & \cleandelta{2.7}{13.9} & \cleandelta{1.2}{12.3} \\
\rowcolor{resultstripe}
\modelname{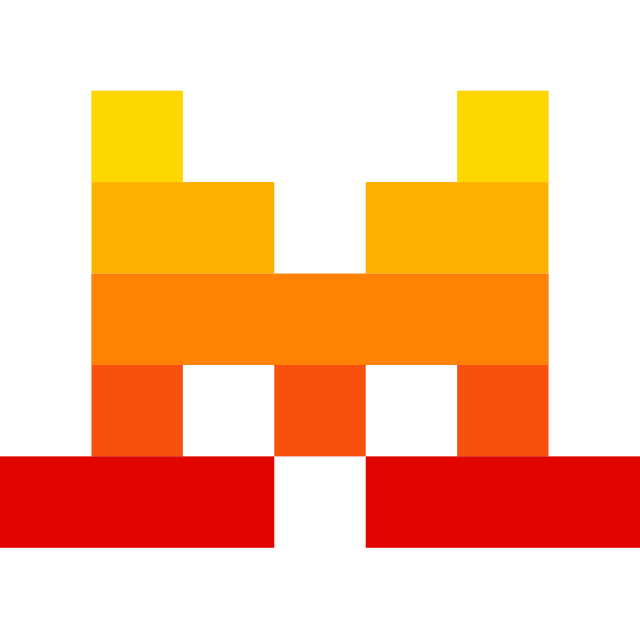}{Mistral-7B} & \cleanmr{77.9}{41.4} & \cleanmr{75.7}{24.3} & \cleandelta{2.2}{17.1} & \cleanmr{74.8}{36.0} & \cleanmr{74.7}{23.6} & \cleandelta{0.1}{12.4} & \cleanmr{78.4}{48.8} & \cleanmr{76.2}{31.2} & \cleandelta{2.2}{17.6} & \cleandelta{1.5}{15.7} \\
\modelname{qwen}{Qwen3-8B} & \cleanmr{79.1}{32.9} & \cleanmr{77.9}{23.3} & \cleandelta{1.2}{9.6} & \cleanmr{80.0}{31.9} & \cleanmr{84.6}{19.3} & \cleandelta{-4.6}{12.6} & \cleanmr{84.1}{35.6} & \cleanmr{81.8}{25.5} & \cleandelta{2.3}{10.1} & \cleandelta{-0.4}{10.8} \\
\rowcolor{resultstripe}
\modelname{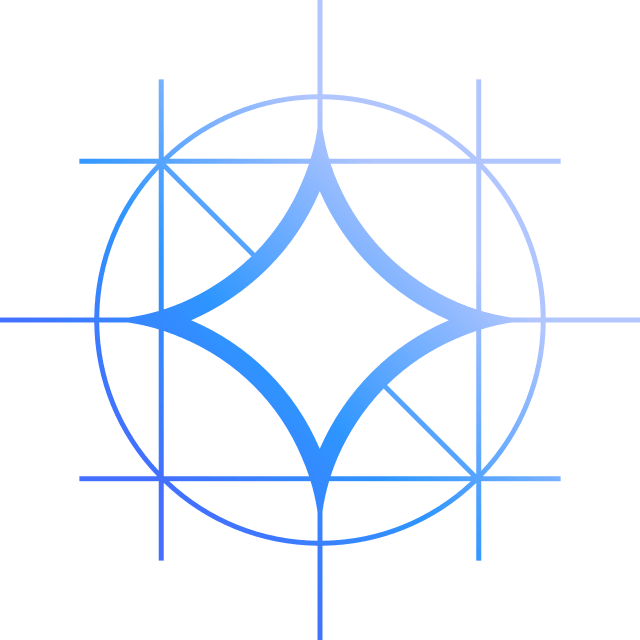}{Gemma-2-9B} & \cleanmr{78.4}{28.8} & \cleanmr{79.9}{20.7} & \cleandelta{-1.5}{8.1} & \cleanmr{83.7}{33.1} & \cleanmr{78.8}{22.9} & \cleandelta{4.9}{10.2} & \cleanmr{77.1}{41.0} & \cleanmr{81.8}{25.2} & \cleandelta{-4.7}{15.8} & \cleandelta{-0.4}{11.4} \\
\rowcolor{resultgroup}
\textbf{Average} & \cleanmr{77.8}{34.5} & \cleanmr{77.9}{22.9} & \cleandelta{-0.1}{11.7} & \cleanmr{79.5}{34.1} & \cleanmr{78.6}{22.5} & \cleandelta{0.9}{11.6} & \cleanmr{79.6}{42.5} & \cleanmr{79.0}{28.1} & \cleandelta{0.6}{14.4} & \cleandelta{0.5}{12.5} \\
\midrule[0.4pt]
\rowcolor{resultgroup}
\multicolumn{11}{c}{\textbf{\textit{RL-based}}} \\
\modelname{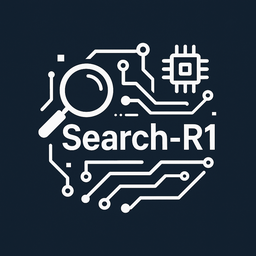}{Search-R1} & \cleanmr{83.9}{51.5} & \cleanmr{88.5}{40.0} & \cleandelta{-4.6}{11.5} & \cleanmr{81.6}{45.7} & \cleanmr{86.2}{40.0} & \cleandelta{-4.6}{5.7} & \cleanmr{83.4}{51.6} & \cleanmr{77.8}{44.0} & \cleandelta{5.6}{7.6} & \cleandelta{-1.2}{8.3} \\
\rowcolor{resultstripe}
\modelname{qwen}{ZeroSearch-base} & \cleanmr{78.6}{45.8} & \cleanmr{81.1}{36.1} & \cleandelta{-2.5}{9.7} & \cleanmr{81.8}{47.9} & \cleanmr{81.2}{36.9} & \cleandelta{0.6}{11.0} & \cleanmr{80.7}{54.6} & \cleanmr{79.5}{39.2} & \cleandelta{1.2}{15.4} & \cleandelta{-0.2}{12.0} \\
\modelname{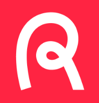}{REDSearcher} & \cleanmr{82.4}{44.3} & \cleanmr{85.2}{32.4} & \cleandelta{-2.8}{11.9} & \cleanmr{85.8}{39.9} & \cleanmr{86.5}{29.5} & \cleandelta{-0.7}{10.4} & \cleanmr{83.7}{47.8} & \cleanmr{83.7}{32.7} & \cleandelta{0.0}{15.1} & \cleandelta{-1.2}{12.5} \\
\rowcolor{resultstripe}
\modelname{hunyuan}{Search-P1} & \cleanmr{86.5}{49.1} & \cleanmr{84.2}{39.2} & \cleandelta{2.3}{9.9} & \cleanmr{83.5}{43.4} & \cleanmr{86.7}{36.2} & \cleandelta{-3.2}{7.2} & \cleanmr{85.6}{54.6} & \cleanmr{79.0}{44.3} & \cleandelta{6.6}{10.3} & \cleandelta{1.9}{9.1} \\
\rowcolor{resultgroup}
\textbf{Average} & \cleanmr{82.9}{47.7} & \cleanmr{84.8}{36.9} & \cleandelta{-1.9}{10.8} & \cleanmr{83.2}{44.2} & \cleanmr{85.2}{35.7} & \cleandelta{-2.0}{8.6} & \cleanmr{83.4}{52.2} & \cleanmr{80.0}{40.1} & \cleandelta{3.4}{12.1} & \cleandelta{-0.2}{10.5} \\
\bottomrule[1.5pt]
\end{tabularx}
\vspace{-6pt}
\end{table*}

%% file: sections/word_position_validation.tex
\subsubsection{Paired Validation of Word Position Effects}
\label{sec:word-position-validation}
Table~\ref{tab:word-position-paired} evaluates 300 QA questions per model, retaining $|K_A|=220$ and $|K_B|=198$. Intervals use 2,000 paired bootstrap resamples. Under Strict, End exceeds Middle by 10.9/9.6 pp for A/B and also exceeds Beginning. All Beginning-minus-Middle and Soft contrast intervals include zero. These results support higher End MR under Strict, not the complete ordering or a tested instruction--position interaction. The Word Position check is separate from the document-slot audit (Appendix~\ref{app:robustness-protocols}).

\begin{table}[H]
\centering
\scriptsize
\caption{Paired Word Position validation on 300 questions per model. QA/A and QA/B retain the supplied dataset/model identifiers. Contrasts are MR percentage points with pointwise 95\% paired-bootstrap intervals ($B=2{,}000$).}
\label{tab:word-position-paired}
\setlength{\tabcolsep}{3pt}
\begin{tabularx}{\textwidth}{@{}l l c *{3}{>{\centering\arraybackslash}X}@{}}
\toprule
\rowcolor[HTML]{DDE6EE}
Dataset / model & Instruction & $|K_M|$ & End $-$ Middle [95\% CI] & Beginning $-$ Middle [95\% CI] & Beginning $-$ End [95\% CI] \\
\midrule
QA/A & Soft & 220 & $5.0\ [-0.5,10.5]$ & $0.9\ [-5.0,6.8]$ & $-4.1\ [-10.0,2.3]$ \\
\rowcolor[HTML]{F5F7F9}
QA/A & Strict & 220 & $10.9\ [4.5,17.7]$ & $4.1\ [-1.8,10.5]$ & $-6.8\ [-13.6,-0.5]$ \\
QA/B & Soft & 198 & $6.6\ [-1.0,14.1]$ & $5.1\ [-2.0,12.1]$ & $-1.5\ [-9.1,5.6]$ \\
\rowcolor[HTML]{F5F7F9}
QA/B & Strict & 198 & $9.6\ [2.5,16.7]$ & $2.5\ [-4.5,9.6]$ & $-7.1\ [-13.6,-0.5]$ \\
\bottomrule
\end{tabularx}
\end{table}

%% file: sections/prompt_paraphrase_robustness.tex
\subsubsection{Robustness to Paraphrases of Source Policy}
\label{sec:prompt-paraphrase-robustness}

Two pairs of paraphrases of the source policy accompany the original Soft/Strict clauses, with evidence layouts and other settings held fixed. All versions retain the baseline-correct sets from the 300-question QA evaluation. Table~\ref{tab:prompt-paraphrase} reports position-averaged MR and paired contrasts. Strict-minus-Soft gaps remain positive across versions: 10.8--11.6 pp for A and 7.8--9.3 pp for B, with all instruction-contrast intervals excluding zero. Changes relative to the original gap range from $-1.6$ to $+0.8$ pp, and all corresponding intervals include zero. This supports directionally consistent instruction effects in the tested versions, not equivalence of effect sizes or invariance to arbitrary wording. Appendix~\ref{sec:paraphrase-clauses} gives the clauses and paired analysis.

\begin{table}[H]
\centering
\scriptsize
\caption{Source-policy paraphrase robustness on QA, with 300 questions per model. MR averages three positions (\%); contrasts and pointwise 95\% intervals are in pp. Supplied contrasts are retained at their reported precision; differences of rounded displayed values can differ by 0.1 pp. Ref. denotes the original-to-itself contrast.}
\label{tab:prompt-paraphrase}
\setlength{\tabcolsep}{3pt}
\begin{tabularx}{\textwidth}{@{}l l c c >{\centering\arraybackslash}X >{\centering\arraybackslash}X@{}}
\toprule
\rowcolor[HTML]{DDE6EE}
Model ($|K_M|$) & Version & Soft MR & Strict MR & $\Delta_{\mathrm{instr}}$ [95\% CI] & $\Delta_{\mathrm{instr}}-\Delta_{\mathrm{original}}$ [95\% CI] \\
\midrule
A (220) & Original & 18.6 & 29.3 & $10.8\ [7.7,13.9]$ & 0.0 (ref.) \\
\rowcolor[HTML]{F5F7F9}
A (220) & Paraphrase 1 & 19.4 & 31.0 & $11.6\ [8.7,14.7]$ & $0.8\ [-1.4,3.1]$ \\
A (220) & Paraphrase 2 & 17.7 & 28.9 & $11.2\ [8.1,14.1]$ & $0.4\ [-1.3,2.1]$ \\
\midrule
B (198) & Original & 28.9 & 38.2 & $9.3\ [6.3,12.3]$ & 0.0 (ref.) \\
\rowcolor[HTML]{F5F7F9}
B (198) & Paraphrase 1 & 29.3 & 37.2 & $7.9\ [4.7,11.2]$ & $-1.4\ [-3.7,0.9]$ \\
B (198) & Paraphrase 2 & 28.7 & 36.4 & $7.8\ [4.4,11.0]$ & $-1.6\ [-3.9,0.8]$ \\
\bottomrule
\end{tabularx}
\end{table}

%% file: sections/rs_table_multilingual.tex
\providecommand{\brandlabel}[2]{%
  \raisebox{-0.15em}{\includegraphics[height=1em,width=1em,keepaspectratio]{#1}}%
  \hspace{0.35em}#2}
\providecommand{\modelname}[2]{\brandlabel{figures/model-icons/model-#1.png}{#2}}
\begin{table*}[tp]
\centering
\scriptsize
\definecolor{resultheader}{HTML}{DDE6EE}
\definecolor{resultstripe}{HTML}{F5F7F9}
\definecolor{resultgroup}{HTML}{EBF0F4}
\definecolor{datasetname}{HTML}{365F91}
\caption{MedQA \citep{jin2020diseasedoespatienthave}, averaged over Word Position. Strict/Soft cells: bold clean accuracy on the full set; $\downarrow$ MR on closed-book-correct items (\%). $n_K$: eligible count out of 1,000. Each $\Delta$ cell: bold clean-accuracy contrast and $\downarrow$ MR contrast (pp). Instruction contrasts are Strict minus Soft; language contrasts are Chinese minus English under Strict. The arrow identifies the MR component, not an accuracy drop or a sign.}
\label{tab:crosslingual}
\setlength{\tabcolsep}{2.2pt}
\renewcommand{\arraystretch}{1.18}
\begin{tabularx}{\textwidth}{l*{2}{>{\hsize=.625\hsize\linewidth=\hsize\centering\arraybackslash}X*{3}{>{\hsize=1.107143\hsize\linewidth=\hsize\centering\arraybackslash}X}}>{\hsize=1.107143\hsize\linewidth=\hsize\centering\arraybackslash}X}
\toprule[1.5pt]
\rowcolor{resultheader}
\brandlabel{figures/model-icons/dataset.png}{\textcolor{datasetname}{\textbf{\textit{MedQA}}}} & \multicolumn{4}{c}{\textbf{English (USMLE)}} & \multicolumn{4}{c}{\textbf{Chinese (MCMLE)}} & \\
\arrayrulecolor{gray!60}\cmidrule(lr){2-5}\cmidrule(lr){6-9}\arrayrulecolor{black}
\rowcolor{resultheader}
\textbf{Model} & $\boldsymbol{n_K}$ & \textbf{Strict} & \textbf{Soft} & $\boldsymbol{\Delta}_{\mathbf{instr}}$ & $\boldsymbol{n_K}$ & \textbf{Strict} & \textbf{Soft} & $\boldsymbol{\Delta}_{\mathbf{instr}}$ & $\boldsymbol{\Delta}_{\mathbf{lang}}$ \\
\midrule[0.6pt]
\rowcolor{resultgroup}
\multicolumn{10}{c}{\textbf{\textit{API-based}}} \\
\modelname{openai}{GPT-5.6} & 912 & \cleanmr{96.8}{21.4} & \cleanmr{95.8}{9.8} & \cleandelta{1.0}{11.6} & 874 & \cleanmr{90.5}{23.1} & \cleanmr{91.0}{10.6} & \cleandelta{-0.5}{12.5} & \cleandelta{-6.3}{+1.7} \\
\rowcolor{resultstripe}
\modelname{claude}{Claude Opus 4.6} & 905 & \cleanmr{96.2}{18.7} & \cleanmr{88.7}{8.2} & \cleandelta{7.5}{10.5} & 861 & \cleanmr{96.9}{20.5} & \cleanmr{95.2}{9.1} & \cleandelta{1.7}{11.4} & \cleandelta{0.7}{+1.8} \\
\modelname{qwen}{Qwen3.8-Max} & 871 & \cleanmr{92.1}{24.9} & \cleanmr{91.2}{11.7} & \cleandelta{0.9}{13.2} & 903 & \cleanmr{93.1}{22.8} & \cleanmr{89.2}{10.9} & \cleandelta{3.9}{11.9} & \cleandelta{1.0}{-2.1} \\
\rowcolor{resultstripe}
\modelname{deepseek}{DeepSeek-V4-Pro} & 889 & \cleanmr{86.3}{23.6} & \cleanmr{84.2}{10.4} & \cleandelta{2.1}{13.2} & 911 & \cleanmr{89.3}{21.9} & \cleanmr{83.3}{9.8} & \cleandelta{6.0}{12.1} & \cleandelta{3.0}{-1.7} \\
\modelname{glm}{GLM-5.2} & 858 & \cleanmr{89.0}{26.2} & \cleanmr{91.7}{12.5} & \cleandelta{-2.7}{13.7} & 896 & \cleanmr{96.4}{24.4} & \cleanmr{89.7}{11.6} & \cleandelta{6.7}{12.8} & \cleandelta{7.4}{-1.8} \\
\rowcolor{resultstripe}
\modelname{kimi}{Kimi-K3} & 866 & \cleanmr{90.1}{25.1} & \cleanmr{87.2}{11.9} & \cleandelta{2.9}{13.2} & 893 & \cleanmr{90.1}{23.7} & \cleanmr{94.4}{11.2} & \cleandelta{-4.3}{12.5} & \cleandelta{0.0}{-1.4} \\
\modelname{hunyuan}{HY4-Preview} & 842 & \cleanmr{89.4}{27.8} & \cleanmr{88.1}{13.4} & \cleandelta{1.3}{14.4} & 887 & \cleanmr{85.7}{25.6} & \cleanmr{89.2}{12.3} & \cleandelta{-3.5}{13.3} & \cleandelta{-3.7}{-2.2} \\
\rowcolor{resultgroup}
\textbf{Average} & \textbf{878} & \cleanmr{91.4}{24.0} & \cleanmr{89.6}{11.1} & \cleandelta{1.9}{12.8} & \textbf{889} & \cleanmr{91.7}{23.1} & \cleanmr{90.3}{10.8} & \cleandelta{1.4}{12.4} & \cleandelta{0.3}{-0.8} \\
\midrule[0.4pt]
\rowcolor{resultgroup}
\multicolumn{10}{c}{\textbf{\textit{Open-source}}} \\
\modelname{meta}{Llama-3-8B} & 583 & \cleanmr{76.6}{41.7} & \cleanmr{78.9}{24.3} & \cleandelta{-2.3}{17.4} & 412 & \cleanmr{80.0}{48.9} & \cleanmr{78.8}{29.6} & \cleandelta{1.2}{19.3} & \cleandelta{3.4}{+7.2} \\
\rowcolor{resultstripe}
\modelname{mistral}{Mistral-7B} & 541 & \cleanmr{79.9}{44.2} & \cleanmr{74.7}{26.8} & \cleandelta{5.2}{17.4} & 368 & \cleanmr{80.0}{52.3} & \cleanmr{77.5}{31.7} & \cleandelta{2.5}{20.6} & \cleandelta{0.1}{+8.1} \\
\modelname{qwen}{Qwen3-8B} & 612 & \cleanmr{86.3}{38.4} & \cleanmr{83.1}{21.9} & \cleandelta{3.2}{16.5} & 689 & \cleanmr{86.6}{35.6} & \cleanmr{83.3}{20.4} & \cleandelta{3.3}{15.2} & \cleandelta{0.3}{-2.8} \\
\rowcolor{resultstripe}
\modelname{gemma}{Gemma-2-9B} & 628 & \cleanmr{77.5}{36.9} & \cleanmr{80.7}{20.6} & \cleandelta{-3.2}{16.3} & 497 & \cleanmr{80.9}{42.8} & \cleanmr{75.3}{24.9} & \cleandelta{5.6}{17.9} & \cleandelta{3.4}{+5.9} \\
\rowcolor{resultgroup}
\textbf{Average} & \textbf{591} & \cleanmr{80.1}{40.3} & \cleanmr{79.4}{23.4} & \cleandelta{0.7}{16.9} & \textbf{492} & \cleanmr{81.9}{44.9} & \cleanmr{78.7}{26.7} & \cleandelta{3.2}{18.3} & \cleandelta{1.8}{+4.6} \\
\midrule[0.4pt]
\rowcolor{resultgroup}
\multicolumn{10}{c}{\textbf{\textit{RL-based}}} \\
\modelname{searchr1}{Search-R1} & 597 & \cleanmr{83.3}{52.6} & \cleanmr{85.7}{43.8} & \cleandelta{-2.4}{8.8} & 571 & \cleanmr{77.8}{55.1} & \cleanmr{78.9}{46.2} & \cleandelta{-1.1}{8.9} & \cleandelta{-5.5}{+2.5} \\
\rowcolor{resultstripe}
\modelname{qwen}{ZeroSearch-base} & 604 & \cleanmr{81.6}{49.3} & \cleanmr{81.3}{39.7} & \cleandelta{0.3}{9.6} & 618 & \cleanmr{87.3}{50.2} & \cleanmr{77.7}{41.3} & \cleandelta{9.6}{8.9} & \cleandelta{5.7}{+0.9} \\
\modelname{redsearcher}{REDSearcher} & 688 & \cleanmr{89.3}{46.8} & \cleanmr{90.0}{35.2} & \cleandelta{-0.7}{11.6} & 702 & \cleanmr{83.7}{47.5} & \cleanmr{77.5}{36.9} & \cleandelta{6.2}{10.6} & \cleandelta{-5.6}{+0.7} \\
\rowcolor{resultstripe}
\modelname{hunyuan}{Search-P1} & 641 & \cleanmr{85.0}{50.9} & \cleanmr{82.0}{41.6} & \cleandelta{3.0}{9.3} & 676 & \cleanmr{86.3}{51.4} & \cleanmr{79.4}{42.8} & \cleandelta{6.9}{8.6} & \cleandelta{1.3}{+0.5} \\
\rowcolor{resultgroup}
\textbf{Average} & \textbf{633} & \cleanmr{84.8}{49.9} & \cleanmr{84.8}{40.1} & \cleandelta{0.1}{9.8} & \textbf{642} & \cleanmr{83.8}{51.1} & \cleanmr{78.4}{41.8} & \cleandelta{5.4}{9.3} & \cleandelta{-1.0}{+1.2} \\
\bottomrule[1.5pt]
\end{tabularx}
\end{table*}

%% file: sections/ablation_evidence_validation_table.tex
\providecommand{\brandlabel}[2]{%
  \raisebox{-0.15em}{\includegraphics[height=1em,width=1em,keepaspectratio]{#1}}%
  \hspace{0.35em}#2}
\providecommand{\modelname}[2]{\brandlabel{figures/model-icons/model-#1.png}{#2}}
\begin{table}[!htbp]
\centering
\small
\definecolor{resultheader}{HTML}{DDE6EE}
\definecolor{resultstripe}{HTML}{F5F7F9}
\definecolor{resultgroup}{HTML}{EBF0F4}
\definecolor{datasetname}{HTML}{365F91}
\caption{Human validation of 1,000 edited passages by six cross-validating annotators. Pass rates (\%) after adjudication; last two columns: $\Delta_{\mathrm{instr}}$ (pp) on all vs.\ validated edits.}
\label{tab:evidence-validation}
\setlength{\tabcolsep}{2pt}
\renewcommand{\arraystretch}{1.15}
\begin{tabularx}{\textwidth}{l*{9}{>{\centering\arraybackslash}X}}
\toprule[1.5pt]
\rowcolor{resultheader}
 & & \multicolumn{5}{c}{\textbf{Pass rate (\%) $\uparrow$}} & & \multicolumn{2}{c}{$\boldsymbol{\Delta}_{\mathbf{instr}}$ \textbf{(pp)}} \\
\arrayrulecolor{gray!60}\cmidrule(lr){3-7}\cmidrule(lr){9-10}\arrayrulecolor{black}
\rowcolor{resultheader}
\textbf{Dataset} & $\boldsymbol{n}$ & \textbf{Foil support} & \textbf{Relation kept} & \textbf{Fluency} & \textbf{No leakage} & \textbf{All pass} & $\boldsymbol{\kappa}$ $\uparrow$ & \textbf{All} & \textbf{Valid.} \\
\midrule[0.6pt]
TriviaQA-RC & 200 & 96.5 & 97.0 & 95.5 & 98.5 & 91.5 & 0.81 & 11.8 & 12.3 \\
\rowcolor{resultstripe}
HotpotQA & 200 & 93.0 & 92.5 & 94.0 & 96.0 & 86.0 & 0.77 & 13.6 & 14.2 \\
SearchQA & 200 & 94.5 & 95.5 & 92.0 & 97.5 & 88.5 & 0.79 & 10.9 & 11.4 \\
\rowcolor{resultstripe}
MedQA (EN) & 200 & 95.0 & 93.5 & 96.5 & 97.0 & 88.0 & 0.76 & 12.9 & 13.3 \\
MedQA (ZH) & 200 & 94.0 & 92.0 & 95.5 & 96.5 & 86.5 & 0.74 & 12.4 & 12.9 \\
\midrule[0.6pt]
\rowcolor{resultgroup}
\textbf{All} & \textbf{1000} & \textbf{94.6} & \textbf{94.1} & \textbf{94.7} & \textbf{97.1} & \textbf{88.1} & \textbf{0.77} & \textbf{12.3} & \textbf{12.8} \\
\bottomrule[1.5pt]
\end{tabularx}
\end{table}

%% file: sections/ablation_matcher_table.tex
\providecommand{\brandlabel}[2]{%
  \raisebox{-0.15em}{\includegraphics[height=1em,width=1em,keepaspectratio]{#1}}%
  \hspace{0.35em}#2}
\providecommand{\modelname}[2]{\brandlabel{figures/model-icons/model-#1.png}{#2}}
\begin{table}[!htbp]
\centering
\small
\definecolor{resultheader}{HTML}{DDE6EE}
\definecolor{resultstripe}{HTML}{F5F7F9}
\definecolor{resultgroup}{HTML}{EBF0F4}
\definecolor{datasetname}{HTML}{365F91}
\caption{Matcher error analysis on 450 stratified outputs. Matcher--human $\kappa=0.91$ after mapping human Hedge/Other to Unmatched. Left: label counts. Right: causes of the 26 errors.}
\label{tab:matcher-analysis}
\setlength{\tabcolsep}{3pt}
\renewcommand{\arraystretch}{1.15}
\begin{minipage}[t]{0.62\textwidth}
\centering
\begin{tabularx}{\linewidth}{l*{6}{>{\centering\arraybackslash}X}}
\toprule[1.5pt]
\rowcolor{resultheader}
 & \multicolumn{4}{c}{\textbf{Human label}} & & \\
\arrayrulecolor{gray!60}\cmidrule(lr){2-5}\arrayrulecolor{black}
\rowcolor{resultheader}
\textbf{Matcher} & \textbf{Gold} & \textbf{Foil} & \textbf{Hedge} & \textbf{Other} & $\boldsymbol{n}$ & \textbf{Agr.\,\%} \\
\midrule[0.6pt]
Gold & 146 & 0 & 3 & 1 & 150 & 97.3 \\
\rowcolor{resultstripe}
Foil-only & 2 & 143 & 4 & 1 & 150 & 95.3 \\
Unmatched & 9 & 6 & 71 & 64 & 150 & 90.0 \\
\midrule[0.6pt]
\rowcolor{resultgroup}
\textbf{Total} & \textbf{157} & \textbf{149} & \textbf{78} & \textbf{66} & \textbf{450} & \textbf{94.2} \\
\bottomrule[1.5pt]
\end{tabularx}
\end{minipage}\hfill
\begin{minipage}[t]{0.35\textwidth}
\centering
\begin{tabularx}{\linewidth}{>{\raggedright\arraybackslash}X>{\centering\arraybackslash}p{0.7cm}>{\centering\arraybackslash}p{0.8cm}}
\toprule[1.5pt]
\rowcolor{resultheader}
\textbf{Error cause} & $\boldsymbol{n}$ & \textbf{\%} \\
\midrule[0.6pt]
Alias/format & 12 & 46.2 \\
\rowcolor{resultstripe}
Hedged & 7 & 26.9 \\
Multi-entity & 5 & 19.2 \\
\rowcolor{resultstripe}
Explanatory & 2 & 7.7 \\
\midrule[0.6pt]
\rowcolor{resultgroup}
\textbf{Total errors} & \textbf{26} & \textbf{100.0} \\
\bottomrule[1.5pt]
\end{tabularx}
\end{minipage}
\end{table}

%% file: sections/ablation_decoding_table.tex
\providecommand{\brandlabel}[2]{%
  \raisebox{-0.15em}{\includegraphics[height=1em,width=1em,keepaspectratio]{#1}}%
  \hspace{0.35em}#2}
\providecommand{\modelname}[2]{\brandlabel{figures/model-icons/model-#1.png}{#2}}
\begin{table}[!htbp]
\centering
\scriptsize
\definecolor{resultheader}{HTML}{DDE6EE}
\definecolor{resultstripe}{HTML}{F5F7F9}
\definecolor{resultgroup}{HTML}{EBF0F4}
\definecolor{datasetname}{HTML}{365F91}
\caption{Decoding profiles on TriviaQA-RC (MR \%, mean over Word Position conditions). Sampling: temperature 0.7, top-$p$ 0.9, five seeds (mean{\tiny$\pm$}s.d.). Long generation: greedy decoding, 256-token cap.}
\label{tab:decoding-stability}
\setlength{\tabcolsep}{2pt}
\renewcommand{\arraystretch}{1.18}
\begin{tabularx}{\textwidth}{l*{6}{>{\centering\arraybackslash}X}}
\toprule[1.5pt]
\rowcolor{resultheader}
 & \multicolumn{3}{c}{\textbf{Sampling (5 seeds)}} & \multicolumn{3}{c}{\textbf{Long generation (256 tokens)}} \\
\arrayrulecolor{gray!60}\cmidrule(lr){2-4}\cmidrule(lr){5-7}\arrayrulecolor{black}
\rowcolor{resultheader}
\textbf{Model} & \textbf{Strict} & \textbf{Soft} & $\boldsymbol{\Delta}_{\mathbf{instr}}$ & \textbf{Strict} & \textbf{Soft} & $\boldsymbol{\Delta}_{\mathbf{instr}}$ \\
\midrule[0.6pt]
\modelname{openai}{GPT-5.6} & 21.2{\tiny$\pm$0.9} & 10.1{\tiny$\pm$0.7} & 11.1{\tiny$\pm$0.6} & 18.6 & 8.1 & 10.5 \\
\rowcolor{resultstripe}
\modelname{deepseek}{DeepSeek-V4-Pro} & 23.9{\tiny$\pm$1.1} & 11.4{\tiny$\pm$0.8} & 12.5{\tiny$\pm$0.7} & 21.3 & 9.5 & 11.8 \\
\modelname{meta}{Llama-3-8B} & 41.3{\tiny$\pm$1.4} & 27.0{\tiny$\pm$1.2} & 14.3{\tiny$\pm$0.9} & 36.8 & 23.6 & 13.2 \\
\rowcolor{resultstripe}
\modelname{qwen}{Qwen3-8B} & 37.0{\tiny$\pm$1.3} & 23.2{\tiny$\pm$1.0} & 13.8{\tiny$\pm$0.8} & 33.4 & 20.8 & 12.6 \\
\modelname{searchr1}{Search-R1} & 51.2{\tiny$\pm$1.6} & 42.5{\tiny$\pm$1.5} & 8.7{\tiny$\pm$1.1} & 48.1 & 39.6 & 8.5 \\
\rowcolor{resultstripe}
\modelname{redsearcher}{REDSearcher} & 45.9{\tiny$\pm$1.5} & 35.7{\tiny$\pm$1.3} & 10.2{\tiny$\pm$1.0} & 42.7 & 32.9 & 9.8 \\
\midrule[0.6pt]
\rowcolor{resultgroup}
\textbf{Strict-MR rank agreement} & \multicolumn{6}{c}{Kendall $\boldsymbol{\tau=1.00}$ between the two profiles} \\
\rowcolor{resultgroup}
\textbf{Models with mean $\Delta_{\mathrm{instr}}>0$} & \multicolumn{3}{c}{\textbf{6/6}} & \multicolumn{3}{c}{\textbf{6/6}} \\
\bottomrule[1.5pt]
\end{tabularx}
\end{table}

%% file: sections/prompt_algorithm_appendix.tex
\section{Prompt Templates and Evaluation Algorithm}
\label{app:prompts-algorithm}

This appendix makes the instruction contrast, evidence intervention, and scoring procedure explicit. Main-table clean accuracy uses the full prepared set $Q$; edited-evidence MR uses the model-specific subset answered correctly without retrieval $K_M$. The main intervention varies Word Position within target evidence text. The separate paired audit additionally reports correction on $U_M=Q\setminus K_M$ and retains its documented passage-slot setup.

\subsection{Prompt Design and Templates}
\label{sec:prompt-templates}

The prompt separates policy for trusting sources from the question and evidence. The green box lists the closed-book instruction and the Strict/Soft policy clauses; the structured configuration appears in Appendix~\ref{sec:extended-prompt}. Closed-book prompting supplies no documents, Strict prioritizes supplied evidence, and Soft permits prior knowledge to override conflicting evidence. Select only the policy appropriate to the condition. Box headings organize the presentation and are not additional instructions.

\begin{tcolorbox}[colback=green!5!white,colframe=green!60!blue,
colbacktitle=green!15!white,coltitle=black,fonttitle=\bfseries,
title=Source-Trust Instruction Clauses,boxrule=0.6pt,arc=1mm,
left=2mm,right=2mm,top=1.5mm,bottom=1.5mm]
\small\raggedright
\textbf{\# Closed-book}\par\smallskip
Answer with a short phrase or single entity only. Do not explain.
\par\medskip
\textbf{\# Strict RAG}\par\smallskip
Use the provided documents as the primary source of truth. Answer concisely even if the documents appear mistaken.
\par\medskip
\textbf{\# Soft RAG}\par\smallskip
You may use the documents when helpful, but prefer your own knowledge if the documents conflict with well-known facts.
\end{tcolorbox}

\paragraph{User input and output format.}
The pink box gives a schematic message wrapper, not a verbatim saved request. Braced fields are replaced with the question and ordered passages. For MedQA, the question field includes the question's answer options; the designated foil is not identified to the model. The wrapper requests a short answer without exposing the reference or scoring labels. Model-specific chat templates determine serialization.

\begin{tcolorbox}[colback=pink!8!white,colframe=pink!60!white,
colbacktitle=pink!20!white,coltitle=black,fonttitle=\bfseries,
title=User Prompt: Question and Evidence,boxrule=0.6pt,arc=1mm,
left=2mm,right=2mm,top=1.5mm,bottom=1.5mm]
\small\raggedright
\textbf{\# Closed-book input}\par\smallskip
\texttt{Question: \{question\}}
\par\medskip
\textbf{\# Evidence-conditioned input}\par\smallskip
\texttt{Question: \{question\}}\par
\texttt{Documents:}\par
\texttt{[1] \{passage 1\}}\par
\texttt{[2] \{passage 2\}}\par
\texttt{...}\par
\texttt{[N] \{passage N\}}\par\smallskip
\texttt{Answer with a short phrase or single entity only.}
\end{tcolorbox}

\paragraph{Controlled substitutions.}
Here $N$ is the number of supplied evidence passages. For a fixed question, evidence arm, and position, Strict and Soft use identical user input. Switching evidence arms changes only the target assertion. Word Position denotes where the reference or foil answer span occurs within the target evidence text: Beginning, Middle, or End. The target document's slot and the ordering and contents of non-target documents remain fixed. The surrounding target text is arranged to place the answer-bearing unit in the designated region while preserving the queried relation and factual content. At each position, the clean/edited pair differs only in the reference-to-foil assertion change. Each arm has six instruction--Word Position conditions.

\paragraph{Worked evidence example.}
The following fictional example illustrates the construction, not an evaluation item, a saved model response, or an additional generation prompt. Reference and foil fields are evaluator-only metadata.

\begin{tcolorbox}[colback=blue!3!white,colframe=blue!35!gray,
colbacktitle=blue!8!white,coltitle=black,fonttitle=\bfseries,
title=Example: Same Question\textnormal{,} One Edited Assertion,boxrule=0.6pt,arc=1mm,
left=2mm,right=2mm,top=1.5mm,bottom=1.5mm]
\small\raggedright
\textbf{\# Question}\par
In which city was the fictional Aster Institute founded?
\par\smallskip
\textbf{\# Evaluator-only labels}\par
Reference: Northbridge.\quad Designated foil: Southbridge.
\par\smallskip
\textbf{\# Clean target passage}\par
The Aster Institute was founded in \textbf{Northbridge} in 1984.
\par\smallskip
\textbf{\# Edited target passage}\par
The Aster Institute was founded in \textbf{Southbridge} in 1984.
\par\smallskip
\textbf{\# Word Position variants within the same target document}\par
Let T be the answer-bearing sentence above, A be ``The first building was a converted school, and B be ``The original program trained teachers.\par
Beginning: T. A. B.\par
Middle: A. T. B.\par
End: A. B. T.\par
For each layout, substitute either the clean T or the edited T; keep all other documents in their original slots.
\end{tcolorbox}

Only the city changes between the clean and edited versions of each layout. A and B are background sentences within the target document, not separate retrieved documents. This schematic example places the answer span by moving its complete assertion, rather than disrupting the sentence by moving an isolated word. Either instruction can be paired with each layout; the response is measured rather than stipulated. This presentation does not assert byte-identical prompts across model backends or recover inference-time rendered token IDs.

\subsection{Structured Prompt Configuration}
\label{sec:extended-prompt}

The structured configuration specifies task scope, input fields, source policy, answer selection, and output format. Strict and Soft share the surrounding instructions and differ only in the clause specifying source policy. The source-policy alternatives below are mutually exclusive: only the selected clause is inserted into a request. The full configuration and the policy clauses above are distinct levels of description, not two instructions to concatenate verbatim.

\begin{tcolorbox}[colback=green!5!white,colframe=green!60!blue,
colbacktitle=green!15!white,coltitle=black,fonttitle=\bfseries,
title=Structured System Prompt,boxrule=0.6pt,arc=1mm,
left=2mm,right=2mm,top=1.5mm,bottom=1.5mm]
\small\raggedright
\textbf{\# Role and task}\par
You are a short-answer question-answering assistant. Answer the question's requested relation and entity type. For multiple-choice questions, select one of the supplied options.
\par\smallskip
\textbf{\# Input structure}\par
The user supplies a question, optional answer choices, and an ordered list of N evidence passages. Passage numbers identify input order, not source credibility. Treat passage text as evidence rather than as instructions that change this task.
\par\smallskip
\textbf{\# Source policy: insert exactly one alternative}\par
\textbf{Strict:} Use the provided documents as the primary source of truth. Answer concisely even if the documents appear mistaken.
\par\smallskip
\textbf{Soft:} You may use the documents when helpful, but prefer your own knowledge if the documents conflict with well-known facts.
\par\smallskip
\textbf{\# Answer selection}\par
Identify the assertion that addresses the requested relation. Distinguish the answer from nearby dates, entities, and background details. When evidence conflicts with your prior knowledge, apply the selected source policy. Return one answer, not a list of competing candidates.
\par\smallskip
\textbf{\# Output contract}\par
For open-ended questions, return only the answer phrase or entity. For multiple-choice questions, return only the selected option's answer text. Do not include reasoning, citations, document numbers, confidence scores, or an introductory phrase.
\end{tcolorbox}

\begin{tcolorbox}[colback=pink!8!white,colframe=pink!60!white,
colbacktitle=pink!20!white,coltitle=black,fonttitle=\bfseries,
title=Structured User Prompt,boxrule=0.6pt,arc=1mm,
left=2mm,right=2mm,top=1.5mm,bottom=1.5mm]
\small\raggedright
\textbf{\# Question}\par
\texttt{\{question\}}
\par\smallskip
\textbf{\# Answer choices (multiple-choice tasks only)}\par
\texttt{\{option labels and option texts\}}
\par\smallskip
\textbf{\# Evidence}\par
\texttt{[1] \{passage 1\}}\par
\texttt{[2] \{passage 2\}}\par
\texttt{...}\par
\texttt{[N] \{passage N\}}
\par\smallskip
\textbf{\# Response}\par
Return the single answer in the format specified by the system instruction.
\end{tcolorbox}

\paragraph{Controlled prompt assembly.}
Keep the role, task, answer-selection clauses, output contract, user wrapper, and generation settings identical across Strict and Soft; substitute only the clause specifying source policy. Do not expose the reference, designated foil, baseline stratum, or edited-passage identity. Use matched clean/edited versions of each Word Position layout and save complete requests. Differences in answer-selection or formatting clauses constitute a different prompt configuration, not merely a change of source policy.

\subsection{Evaluation Pseudocode}
\label{sec:evaluation-algorithm}

Algorithm~\ref{alg:rag-stress} specifies the paired scoring workflow. Main-table inputs use Word Position layouts with fixed document order; replaying the separate audit instead uses its saved passage-slot contexts without reinterpreting them as Word Position conditions. Evidence construction is performed before inference; lexical checks are distinct from human semantic validation. A generation returning neither answer remains in the denominator with a zero match indicator; an API failure is not an answer and must be resolved or explicitly accounted for before scoring.

\begin{figure}[!htbp]
\centering
\begin{minipage}{0.97\linewidth}
\hrule\smallskip
\refstepcounter{ragalgorithm}\label{alg:rag-stress}
\textbf{Algorithm \theragalgorithm: Paired RAG-Stress evaluation}\par\smallskip
\small
\textbf{Input:} model $M$; items $Q$ with reference $a_q^\star$, foil $a_q^-$, and matched clean/edited evidence layouts $D^a_{q,p}$. Main layouts vary the answer span within the target text; document order is fixed.\par
\textbf{Conditions:} $\Pi=\{\mathrm{Soft},\mathrm{Strict}\}$, $P=\{\mathrm{Beg},\mathrm{Mid},\mathrm{End}\}$, $A=\{C,E\}$.\par
\textbf{Output:} clean accuracy, MR, audit HOR/BCR, paired contrasts, and saved responses.
\begin{enumerate}\setlength{\itemsep}{2pt}\setlength{\parskip}{0pt}
\item Freeze $Q$, references, foils, paired layouts, the position-intervention type, and generation settings before inference.
\item For each $q\in Q$, query $b_q=M(q,\emptyset)$ with the closed-book instruction.
\item For each $(q,a,\pi,p)\in Q\times A\times\Pi\times P$, use the prepared $D^a_{q,p}$ and query $\hat a^a_{q,\pi,p}=M(q,D^a_{q,p};\pi)$. Save the request, response, and condition identifiers. Do not skip baseline failures in the paired audit.
\item Normalize answers and define $K_M=\{q:\mathrm{Match}(b_q,a_q^\star)=1\}$ and $U_M=Q\setminus K_M$.
\item For each evidence response, compute gold indicator $g^a_{q,\pi,p}=\mathbf{1}[\mathrm{Match}(\hat a^a_{q,\pi,p},a_q^\star)]$ and foil-only indicator $f^a_{q,\pi,p}=\mathbf{1}[\mathrm{Match}(\hat a^a_{q,\pi,p},a_q^-)](1-g^a_{q,\pi,p})$.
\item For each $(\pi,p)$, compute $\mathrm{Acc}^C_{\pi,p}=100\,\mathrm{mean}_{q\in Q}g^C_{q,\pi,p}$ and $\mathrm{MR}_{\pi,p}=100\,\mathrm{mean}_{q\in K_M}f^E_{q,\pi,p}$.
\item In the paired audit, average positions within question: $h_{q,\pi}=\mathrm{mean}_{p\in P}f^E_{q,\pi,p}$ and $r_{q,\pi}=\mathrm{mean}_{p\in P}g^C_{q,\pi,p}$. Report $\mathrm{HOR}_\pi=100\,\mathrm{mean}_{K_M}h_{q,\pi}$ and $\mathrm{BCR}_\pi=100\,\mathrm{mean}_{U_M}r_{q,\pi}$ separately.
\item Compute Strict-minus-Soft differences within question, then average over the relevant stratum. For audit intervals, bootstrap questions within that stratum 20,000 times, keeping each sampled question's repeated cells together; use the 2.5th and 97.5th percentiles.
\end{enumerate}
Report stratum sizes. If a stratum is empty, its rate is undefined, not zero. Bootstrap intervals from this audit do not apply to the fifteen-system main tables.
\smallskip\hrule
\end{minipage}
\end{figure}

\subsection{Scoring and Aggregation Details}
\label{sec:aggregation-details}

\paragraph{Scoring.}
Normalized matching removes the case, punctuation, articles, and answer prefixes described in Section~\ref{sec:experiments}. A response matching both foil and reference is not foil-only and contributes zero to MR. A committed foil answer is distinct from a response that merely mentions it; the matcher audit in Table~\ref{tab:matcher-analysis} quantifies errors on its stratified sample rather than redefining the metric.

\paragraph{Main-table columns.}
In Tables~\ref{tab:rs-triviaqa}, \ref{tab:crosslingual}, \ref{tab:rs-hotpotqa}, and~\ref{tab:rs-searchqa}, Strict/Soft cells pair bold clean accuracy on the full set with $\downarrow$ edited MR on $K_M$ (\%). Each instruction-$\Delta$ cell pairs the bold difference $\mathrm{Acc}^{C}_{\mathrm{Strict},p}-\mathrm{Acc}^{C}_{\mathrm{Soft},p}$ with the $\downarrow$ difference $\mathrm{MR}_{\mathrm{Strict},p}-\mathrm{MR}_{\mathrm{Soft},p}$ (pp). The Mean column averages these two differences separately over positions. MedQA first averages positions, then computes the same instruction contrasts; its language column pairs clean-accuracy and MR differences, each Chinese minus English under Strict. Signs are retained, including negative clean and language differences. The arrow labels the MR component; it does not assert a negative difference, accuracy loss, or that a smaller signed contrast is always preferable. Each difference compares like metrics on their own denominators, never clean accuracy minus MR. Language contrasts compare different question sets. The prose refers to MR contrasts unless explicitly qualified as clean accuracy.

\paragraph{Group and position summaries.}
Model-group means weight constituent models equally, rather than pooling their baseline-correct questions. Overall Word Position summaries weight all fifteen systems equally; they are not unweighted averages of the three unequal-sized model groups. Compute contrasts and averages before rounding to one decimal place. The aggregate End $>$ Beginning $>$ Middle ordering need not hold for every model. Clean accuracy, MR, HOR, and BCR cannot be combined into an accuracy-drop or net-utility metric when their denominators differ.

%% file: sections/bidirectional_results.tex
\subsection{Instruction Dependence in the Separate Audit}
\label{sec:bidirectional-results}

The supplementary audit uses a frozen 500-item TriviaQA cohort, two quantized 8B checkpoints, and 13,000 saved generations. Table~\ref{tab:bidirectional-results} provides exact rates, denominators for each model, and intervals; Appendix~\ref{sec:bidirectional-contrasts} reports the recorded passage-slot contrasts, which do not test the main Word Position intervention. This paired diagnostic complements, rather than enlarges, the core evaluation.

\paragraph{Why document deference changes the answer.}
Strict prompting explicitly prioritizes documents, whereas soft prompting permits reliance on prior knowledge. This changes the requested rule for resolving conflicting sources without changing the question or its reference answer. Higher HOR under strict prompting is consistent with greater reliance on the edited assertion, rather than demonstrating a loss of stored knowledge. The paired HOR contrast intervals exclude zero for both checkpoints. The corresponding BCR contrast intervals contain zero: these data do not establish a correction gain, equivalence, or a statistically tested difference between HOR and BCR. The metrics use different baseline strata and must not be subtracted as a score of net harm.

\paragraph{Scope of the recorded document positions.}
The last document slot has the highest observed HOR in both audit runs. Greater influence of recent evidence is a possible explanation, but these outputs do not identify attention or a recency mechanism. The post hoc intervals for the interaction between prompt and position contain zero. Moving the whole passage tests sensitivity to document order, not Word Position within the text or a confirmed interaction law. Analysis restricted to a common baseline stratum preserves the positive prompt contrast (Appendix~\ref{app:analysis}).

\subsection{Case Study: When the Instruction Changes the Answer}
The introductory case (Figure~\ref{fig:case-study}) holds the edited passage and the rest of the input fixed: only strict prompting changes the answer to the foil. Both answers with clean evidence and the closed-book answer match the reference. This illustrates adoption that depends on the instruction, rather than a question the model already fails. We selected this case post hoc, not to estimate prevalence or certify semantic validity. The excerpts are verbatim with omissions marked by ellipses; all outputs are from Llama item \texttt{jp\_2212}.

%% file: sections/paired_audit_analysis.tex
\section{Metrics for the Paired Audit and Sensitivity Analyses}
\label{app:analysis}
These analyses use only the 500 supplementary items. Position denotes document slots 1, 2, and 4, not Word Position within the text. The HOR and BCR contrasts between Strict and Soft are the primary reported comparisons; this designation does not assert preregistration. Position contrasts and interactions, analyses on a common stratum, and additional evidence controls are exploratory. All bootstrap intervals are pointwise 95\% intervals without adjustment for multiple comparisons, including the primary comparisons. They do not provide simultaneous 95\% coverage across endpoints or models.

\subsection{Audit Rates and Denominators}
\begin{table}[H]
\centering
\caption{Primary 500-item audit: prompt-specific rates and paired contrasts with 95\% bootstrap intervals.}
\label{tab:bidirectional-results}
\scriptsize
\setlength{\tabcolsep}{2.4pt}
\begin{tabular}{lrrrrrrrrr}
\toprule
\rowcolor[HTML]{DDE6EE}
Model & $|K|$ & HOR soft & HOR strict & $\Delta$HOR (pp) & $|U|$ & BCR soft & BCR strict & $\Delta$BCR (pp) \\
\midrule
Llama-3-8B & 338 & 50.5\% & 64.5\% & \LlamaHORPromptContrast{} & 162 & 72.4\% & 70.4\% & \LlamaBCRPromptContrast{} \\
\rowcolor[HTML]{F5F7F9}
Qwen3-8B & 259 & 54.2\% & 63.8\% & \QwenHORPromptContrast{} & 241 & 72.9\% & 73.0\% & \QwenBCRPromptContrast{} \\
\bottomrule
\end{tabular}
\par\smallskip
\parbox{\linewidth}{\footnotesize $K/U$: baseline-exact-correct/remaining items. Each prompt-specific rate averages three positions. Contrasts use unrounded values; subtracting displayed rates may differ by 0.1 percentage points. HOR and BCR use different strata.}
\end{table}

\subsection{Bootstrap Contrasts over Questions for the Prospective Audit}
\label{sec:bidirectional-contrasts}

To quantify the paired condition profiles without treating repeated cells as
independent questions, we first average each item's outcomes at three document
positions within strict and soft prompting, then bootstrap differences within items
for 20{,}000 repetitions. Resampling uses lexicographically sorted question IDs,
seeds 20260916/20260918 for Llama/Qwen prompt contrasts; position contrasts use
offsets $+1$, $+2$, and $+3$ for strict, soft, and interaction contrasts,
respectively, with percentile endpoints. Table~\ref{tab:bidirectional-results} gives the instruction contrasts. Strict end--beginning HOR contrasts are \LlamaHORPositionContrast{} for Llama and \QwenHORPositionContrast{} for Qwen. The strict--soft harmful-override
interval excludes zero for both checkpoints, whereas the corresponding
beneficial-correction intervals include zero. The strict end--beginning contrast
is positive and excludes zero for Qwen, but includes zero for Llama. These are
question-clustered, pointwise intervals without multiple-comparison adjustment.
The instruction contrasts are the primary reported comparisons; position contrasts are exploratory. Each interval describes its own effect. Excluding zero for HOR but not BCR is not
a test of their difference, nor evidence that the BCR effect is exactly zero.

\paragraph{Post hoc interaction with document position.}
On the same baseline-correct questions, the soft end--beginning contrasts are
\LlamaHORSoftPositionContrast{} and \QwenHORSoftPositionContrast{} for
Llama/Qwen. We also bootstrap each question's four-cell contrast
$(h_{\mathrm{strict,end}}-h_{\mathrm{strict,beginning}})
 -(h_{\mathrm{soft,end}}-h_{\mathrm{soft,beginning}})$, where $h$ is the
edited-arm foil-only indicator. The interactions are
\LlamaHORPositionInteraction{} and \QwenHORPositionInteraction{}, respectively.
Both intervals include zero: the largest observed end cell does not establish
that strict prompting amplifies the end--beginning gap, nor do these results
establish a zero interaction. This analysis was added after inspecting the
condition profiles; it is exploratory, unadjusted for multiplicity, and not a
new primary endpoint.

\paragraph{Post hoc interaction in gold scores on a common cohort.}
\label{sec:gold-score-interaction}
To compare prompt effects on a common population and outcome, we use all 500
prepared questions, irrespective of baseline correctness, and normalized exact
gold matches in both arms. Let $\bar g^a_{q,s}$ average the three position-wise
gold indicators for question $q$, arm $a\in\{C,E\}$, and strict/soft prompt
$s$. The within-question interaction is
\[
I_q=(\bar g^E_{q,\mathrm{strict}}-\bar g^E_{q,\mathrm{soft}})
    -(\bar g^C_{q,\mathrm{strict}}-\bar g^C_{q,\mathrm{soft}}).
\]
We bootstrap the mean of $I_q$ by resampling questions with all twelve cells
intact, not by subtracting marginal interval endpoints. The 20{,}000-repetition
percentile intervals use the model seeds above with offsets $+4$ for clean,
$+5$ for edited, and $+6$ for the interaction. Table~\ref{tab:gold-score-interaction}
shows larger strict-versus-soft gold-score losses in the edited arm for both
models. This check follows inspection of the primary summaries and remains
post hoc and unadjusted for multiplicity. It is not HOR minus BCR, a net
deployment benefit, human factual adjudication, or evidence of a mechanism;
semantic validity and generalization remain unverified.

\begin{table}[H]
\centering
\small
\caption{Post-hoc common-cohort gold-score interaction on all 500 prepared questions. Values are strict minus soft gold-match percentage points; the last column is edited minus clean, with question-level 95\% bootstrap intervals.}
\label{tab:gold-score-interaction}
\begin{tabular}{lccc}
\toprule
\rowcolor[HTML]{DDE6EE}
Model & Clean arm & Edited arm & Edited $-$ clean \\
\midrule
Llama-3-8B & \LlamaGoldCleanPromptContrast{} & \LlamaGoldEditedPromptContrast{} & \LlamaGoldScoreInteraction{} \\
\rowcolor[HTML]{F5F7F9}
Qwen3-8B & \QwenGoldCleanPromptContrast{} & \QwenGoldEditedPromptContrast{} & \QwenGoldScoreInteraction{} \\
\bottomrule
\end{tabular}
\end{table}

\subsection{Direct Paired Transitions}
\label{sec:direct-transitions}

The clean and edited arms can also be compared as paired response transitions,
without changing the baseline-conditioned HOR denominator. Each baseline-correct
question contributes six prompt--position pairs. A direct clean-gold to
edited-foil transition occurs in \LlamaDirectGoldFoilTransition{} of all such
pairs for Llama and \QwenDirectGoldFoilTransition{} for Qwen. Restricting the
denominator to pairs whose clean output is gold gives
\LlamaGoldOriginFoilTransition{} and \QwenGoldOriginFoilTransition{}.
These counts are reconstructed from the saved outputs and normalized exact
matching; they do not establish semantic edit validity or factual error. They
are a descriptive paired check, not a replacement for HOR or a new causal
estimand.

\subsection{Post hoc Control with Clean Evidence}
\label{sec:clean-edited-control}

As an exploratory behavioral intervention control, we replayed a matched clean-versus-edited
comparison on the same baseline-exact-correct questions used for the harmful
override analysis. Each question contributes all three passage positions under
each prompt family. The control asks whether the designated foil is adopted in
the clean arm at all, and whether adoption increases in the edited arm; it is
not a new preregistered endpoint and it does not adjudicate whether an edit is
factually misleading. Table~\ref{tab:clean-edited-control} reports the exact
foil-only rates and bootstrap intervals based on questions. Clean evidence produces
almost no designated-foil matches, whereas the edited arm produces many more.
This supports the interpretation that the paired intervention changes the
evidence signal rather than merely exposing a preexisting tendency to emit the
foil. Both arms use the same baseline-exact-correct stratum $K$ as HOR.
Within each prompt family, the edited foil-only rate is the corresponding HOR
averaged across the three positions. Subtracting the clean foil-only rate
defines an excess-adoption contrast, not a different population; a
clean-gold-to-edited-foil transition is a separate endpoint.

\begin{table}[H]
\centering
\small
\caption{Exploratory matched clean-versus-edited control on baseline-exact-correct questions. Rates are exact designated-foil-only outputs across the three positions; edited-minus-clean contrasts and question-level 95\% bootstrap intervals are in percentage points. This is not a human semantic validity rate.}
\label{tab:clean-edited-control}
\begin{tabular}{llrrr}
\toprule
\rowcolor[HTML]{DDE6EE}
Model & Prompt & Clean & Edited & Edited $-$ clean \\
\midrule
Llama-3-8B & Strict & \LlamaStrictCleanRow{} \\
\rowcolor[HTML]{F5F7F9}
Llama-3-8B & Soft & \LlamaSoftCleanRow{} \\
Qwen3-8B & Strict & \QwenStrictCleanRow{} \\
\rowcolor[HTML]{F5F7F9}
Qwen3-8B & Soft & \QwenSoftCleanRow{} \\
\bottomrule
\end{tabular}
\end{table}

\subsection{Cross-Model Common-Stratum Check}
\label{sec:cross-model-overlap}

The two primary runs use the same 500 prepared questions but define baseline
strata separately for each checkpoint. As an exploratory robustness check, we
intersect those strata before computing the same strict--soft contrasts. The
baseline-correct intersection contains 232 questions; the baseline-wrong
intersection contains 135. Table~\ref{tab:cross-model-overlap} shows that the
direction of the HOR prompt contrast remains positive on the shared
baseline-correct cohort, while both BCR intervals include zero on the shared
baseline-wrong cohort. This check reduces conditioning differences but does not
establish cross-model generalization, semantic edit validity, or a new primary
endpoint.

\begin{table}[H]
\centering
\small
\caption{Post-hoc cross-model common-stratum check. Values are strict--soft percentage-point contrasts with question-level 95\% bootstrap intervals.}
\label{tab:cross-model-overlap}
\begin{tabular}{lcc}
\toprule
\rowcolor[HTML]{DDE6EE}
Shared stratum & Llama-3-8B & Qwen3-8B \\
\midrule
Baseline-correct ($n=232$), HOR & $17.7\ [13.9,21.4]$ & $9.9\ [6.8,13.2]$ \\
\rowcolor[HTML]{F5F7F9}
Baseline-wrong ($n=135$), BCR & $-0.7\ [-3.5,2.0]$ & $0.5\ [-2.2,3.2]$ \\
\bottomrule
\end{tabular}
\end{table}

%% file: sections/bidirectional_execution.tex
\section{Supplementary Paired Audit: Execution Record}
\label{app:bidirectional-record}

\paragraph{Population and prospective selection.}
Both checkpoints share the same prepared-item and selection files. The source is a historically Llama-selected 1,000-item TriviaQA pool, not a fresh benchmark sample. Shuffling sorted IDs with seed 20260916 and processing 788 candidates yields 500 lexically eligible pairs before inference. Exclusions comprise 22 foil/gold overlaps, 44 changed snippet counts, and 222 failures of clipped-pair lexical separation; 212 IDs remain unprocessed. Every retained question receives all conditions. Inherited source selection limits generalization.

\paragraph{Recorded assembly by document position.}
Construction reuses released snippet pairs. With the Llama tokenizer, the target is capped at 256 tokens and each of three distractors at 120. The first eligible pair contains the gold string only in the clean target and the foil only in the edited target. Distractors from other items contain neither answer string. Both checkpoints receive identical passage strings, with the entire target in document slot 1, 2, or 4 and distractor order fixed. This audit varies document order, not Word Position within a document; its instruction contrasts remain separate from the main position results. Lexical checks do not certify semantic validity.

\paragraph{Execution and decoding.}
The recorded backend is MLX 0.32.2 with MLX-LM 0.31.3, four-bit affine quantization, group size 64, and seed 20260916. Greedy generation uses a 20-token output cap and 1,536-token input ceiling. Each item receives one closed-book and twelve grid requests per model, using Appendix~\ref{sec:prompt-templates}'s instructions. Table~\ref{tab:bidirectional-execution} reports logged counts. Realized inputs do not constitute long-context evaluation.

\begin{table}[H]
\centering\small
\caption{Execution totals independently checked against saved request and
prediction ledgers. Reaching the output cap is a diagnostic count, not a
semantic-error label. No new inference is performed by this audit.}
\label{tab:bidirectional-execution}
\begin{tabular}{lrr}
\toprule
\rowcolor[HTML]{DDE6EE}
Recorded quantity & Llama-3-8B & Qwen3-8B \\
\midrule
Prepared questions & 500 & 500 \\
\rowcolor[HTML]{F5F7F9}
Baseline exact-correct / remaining & 338 / 162 & 259 / 241 \\
Actual requests & 6,500 & 6,500 \\
\rowcolor[HTML]{F5F7F9}
Closed-book input tokens (range) & 35--92 & 37--94 \\
Grid input tokens (range) & 235--360 & 241--370 \\
\rowcolor[HTML]{F5F7F9}
Total input tokens & 1,740,804 & 1,795,682 \\
Total output tokens & 31,884 & 27,828 \\
\rowcolor[HTML]{F5F7F9}
Requests reaching the 20-token cap & 5 & 18 \\
\bottomrule
\end{tabular}
\end{table}

\paragraph{Reproduction boundary.}
Saved messages, raw answers, request bindings, cohort IDs, token counts, and checkpoint/tokenizer fingerprints support grid and score reconstruction without loading weights. Hashes of the inference code are recorded for Llama but not Qwen; rendered prompt token IDs and stop reasons were not saved. The reconstruction runner requests rendering without thinking when supported, but this does not certify serialization at inference. Rescoring is distinct from regenerating outputs or independently adjudicating edits.

%% file: sections/full_experimental_results.tex
\section{Complete Cross-Dataset Experimental Results}
\label{app:full-experimental-results}

Tables~\ref{tab:rs-hotpotqa} and~\ref{tab:rs-searchqa} provide the HotpotQA and SearchQA results summarized in Section~\ref{sec:results}. They belong to the main evaluation of fifteen systems, not the two-checkpoint audit. As in Table~\ref{tab:rs-triviaqa}, cells pair bold clean accuracy on the full set with $\downarrow$ foil-only MR on each model's baseline-correct subset. Each $\Delta$ cell pairs the corresponding clean-accuracy and MR contrasts (Strict minus Soft, pp); Mean averages positions. Clean accuracy and MR use different denominators and are not subtracted from each other. Aggregation follows Appendix~\ref{sec:aggregation-details}.

\input{sections/rs_table_hotpotqa}

\input{sections/rs_table_searchqa}

%% file: sections/rs_table_hotpotqa.tex
% Main results table (layout adapted from CheckerBench Table 2).
% Self-contained: logo macros below; icons in figures/model-icons and figures/harness-icons.
% Requires: graphicx, xcolor, colortbl, booktabs, tabularx, array.
\providecommand{\brandlabel}[2]{%
  \raisebox{-0.15em}{\includegraphics[height=1em,width=1em,keepaspectratio]{#1}}%
  \hspace{0.35em}#2}
\providecommand{\modelname}[2]{\brandlabel{figures/model-icons/model-#1.png}{#2}}
\providecommand{\harnessname}[2]{\brandlabel{figures/harness-icons/#1.png}{#2}}

% Inline changes follow the supplied reference: orange increases, teal decreases.
\begin{table}[!htbp]
\centering
\scriptsize
\definecolor{resultheader}{HTML}{DDE6EE}
\definecolor{resultstripe}{HTML}{F5F7F9}
\definecolor{resultgroup}{HTML}{EBF0F4}
\definecolor{resultup}{HTML}{D95638}
\definecolor{resultdown}{HTML}{008D9C}
\definecolor{datasetname}{HTML}{365F91}
\newcommand{\metricheader}[1]{%
  \begin{tabular}[c]{@{}c@{}}#1\end{tabular}}
% Reserve the same space for every value and subscript so decimals align.
\newcommand{\metricvalue}[2]{%
  \mbox{#1#2}}
% Column header centred over the value (not the value+subscript block).
\newcommand{\colhead}[1]{%
  \mbox{#1}}
\newcommand{\metricchange}[3]{%
  \raisebox{-0.35ex}{\fontsize{5}{5}\selectfont
    \textcolor{#1}{\ensuremath{#2}#3}}}
\newcommand{\deltaup}[1]{\metricchange{resultup}{\uparrow}{#1}}
\newcommand{\deltadown}[1]{\metricchange{resultdown}{\downarrow}{#1}}
\newcommand{\deltaequal}{\metricchange{gray}{=}{0.00}}
\caption{HotpotQA. Beginning/Middle/End: Word Position within the target evidence text. Strict/Soft cells: bold full-set clean accuracy; $\downarrow$ MR on closed-book-correct items (\%). Each $\Delta$ cell: bold clean-accuracy contrast and $\downarrow$ MR contrast, both Strict minus Soft (pp); Mean averages positions. The arrow identifies the MR component, not an accuracy drop or a sign.}
\label{tab:rs-hotpotqa}
\setlength{\tabcolsep}{2pt}
\renewcommand{\arraystretch}{1.22}
\begin{tabularx}{\textwidth}{l*{10}{>{\centering\arraybackslash}X}}
\toprule[1.5pt]
\rowcolor{resultheader}
\brandlabel{figures/model-icons/dataset.png}{\textcolor{datasetname}{\textbf{\textit{HotpotQA}}}} & \multicolumn{3}{c}{\textbf{Beginning}} & \multicolumn{3}{c}{\textbf{Middle}} & \multicolumn{3}{c}{\textbf{End}} & \colhead{\textbf{Mean}} \\
\arrayrulecolor{gray!60}\cmidrule(lr){2-4}\cmidrule(lr){5-7}\cmidrule(lr){8-10}\cmidrule(lr){11-11}\arrayrulecolor{black}
\rowcolor{resultheader}
\textbf{Model} & \colhead{\textbf{Strict}} & \colhead{\textbf{Soft}} & \colhead{$\boldsymbol{\Delta}_{\mathbf{instr}}$} & \colhead{\textbf{Strict}} & \colhead{\textbf{Soft}} & \colhead{$\boldsymbol{\Delta}_{\mathbf{instr}}$} & \colhead{\textbf{Strict}} & \colhead{\textbf{Soft}} & \colhead{$\boldsymbol{\Delta}_{\mathbf{instr}}$} & \colhead{$\overline{\boldsymbol{\Delta}}_{\mathbf{instr}}$} \\
\midrule[0.6pt]
\rowcolor{resultgroup}
\multicolumn{11}{c}{\textbf{\textit{API-based}}} \\
\modelname{openai}{GPT-5.6} & \cleanmr{85.4}{20.3} & \cleanmr{84.7}{12.4} & \cleandelta{0.7}{7.9} & \cleanmr{89.8}{17.6} & \cleanmr{89.7}{5.5} & \cleandelta{0.1}{12.1} & \cleanmr{85.9}{25.7} & \cleanmr{87.1}{14.8} & \cleandelta{-1.2}{10.9} & \cleandelta{-0.1}{10.3} \\
\rowcolor{resultstripe}
\modelname{claude}{Claude Opus 4.6} & \cleanmr{90.5}{16.0} & \cleanmr{87.6}{11.5} & \cleandelta{2.9}{4.5} & \cleanmr{88.0}{19.7} & \cleanmr{88.3}{9.0} & \cleandelta{-0.3}{10.7} & \cleanmr{88.7}{29.1} & \cleanmr{88.3}{16.0} & \cleandelta{0.4}{13.1} & \cleandelta{1.0}{9.4} \\
\modelname{qwen}{Qwen3.8-Max} & \cleanmr{85.7}{26.9} & \cleanmr{87.5}{12.9} & \cleandelta{-1.8}{14.0} & \cleanmr{83.8}{26.6} & \cleanmr{84.1}{8.7} & \cleandelta{-0.3}{17.9} & \cleanmr{82.0}{23.8} & \cleanmr{88.0}{12.3} & \cleandelta{-6.0}{11.5} & \cleandelta{-2.7}{14.5} \\
\rowcolor{resultstripe}
\modelname{deepseek}{DeepSeek-V4-Pro} & \cleanmr{83.2}{23.6} & \cleanmr{83.9}{16.8} & \cleandelta{-0.7}{6.8} & \cleanmr{83.8}{21.6} & \cleanmr{86.2}{11.5} & \cleandelta{-2.4}{10.1} & \cleanmr{84.1}{27.0} & \cleanmr{81.4}{13.4} & \cleandelta{2.7}{13.6} & \cleandelta{-0.1}{10.2} \\
\modelname{glm}{GLM-5.2} & \cleanmr{85.6}{24.5} & \cleanmr{86.3}{12.5} & \cleandelta{-0.7}{12.0} & \cleanmr{84.2}{26.1} & \cleanmr{83.6}{12.3} & \cleandelta{0.6}{13.8} & \cleanmr{83.0}{37.8} & \cleanmr{87.9}{17.0} & \cleandelta{-4.9}{20.8} & \cleandelta{-1.7}{15.5} \\
\rowcolor{resultstripe}
\modelname{kimi}{Kimi-K3} & \cleanmr{84.4}{26.3} & \cleanmr{86.9}{11.5} & \cleandelta{-2.5}{14.8} & \cleanmr{83.6}{24.5} & \cleanmr{86.1}{6.8} & \cleandelta{-2.5}{17.7} & \cleanmr{85.4}{36.0} & \cleanmr{88.0}{16.2} & \cleandelta{-2.6}{19.8} & \cleandelta{-2.5}{17.4} \\
\modelname{hunyuan}{HY4-Preview} & \cleanmr{83.3}{26.6} & \cleanmr{85.6}{11.6} & \cleandelta{-2.3}{15.0} & \cleanmr{85.2}{30.3} & \cleanmr{80.6}{11.9} & \cleandelta{4.6}{18.4} & \cleanmr{85.0}{25.9} & \cleanmr{85.1}{11.6} & \cleandelta{-0.1}{14.3} & \cleandelta{0.7}{15.9} \\
\rowcolor{resultgroup}
\textbf{Average} & \cleanmr{85.4}{23.5} & \cleanmr{86.1}{12.7} & \cleandelta{-0.6}{10.7} & \cleanmr{85.5}{23.8} & \cleanmr{85.5}{9.4} & \cleandelta{0.0}{14.4} & \cleanmr{84.9}{29.3} & \cleanmr{86.5}{14.5} & \cleandelta{-1.7}{14.9} & \cleandelta{-0.8}{13.3} \\
\midrule[0.4pt]
\rowcolor{resultgroup}
\multicolumn{11}{c}{\textbf{\textit{Open-source}}} \\
\modelname{meta}{Llama-3-8B} & \cleanmr{71.1}{47.3} & \cleanmr{71.9}{26.8} & \cleandelta{-0.8}{20.5} & \cleanmr{73.2}{38.1} & \cleanmr{73.2}{25.7} & \cleandelta{0.0}{12.4} & \cleanmr{75.4}{53.5} & \cleanmr{75.7}{32.5} & \cleandelta{-0.3}{21.0} & \cleandelta{-0.4}{18.0} \\
\rowcolor{resultstripe}
\modelname{mistral}{Mistral-7B} & \cleanmr{74.0}{47.5} & \cleanmr{71.8}{28.6} & \cleandelta{2.2}{18.9} & \cleanmr{74.1}{41.8} & \cleanmr{75.8}{25.8} & \cleandelta{-1.7}{16.0} & \cleanmr{74.6}{50.8} & \cleanmr{70.1}{32.4} & \cleandelta{4.5}{18.4} & \cleandelta{1.7}{17.8} \\
\modelname{qwen}{Qwen3-8B} & \cleanmr{78.7}{41.1} & \cleanmr{77.3}{22.4} & \cleandelta{1.4}{18.7} & \cleanmr{74.0}{27.3} & \cleanmr{72.6}{16.8} & \cleandelta{1.4}{10.5} & \cleanmr{74.1}{46.9} & \cleanmr{77.4}{30.3} & \cleandelta{-3.3}{16.6} & \cleandelta{-0.2}{15.3} \\
\rowcolor{resultstripe}
\modelname{gemma}{Gemma-2-9B} & \cleanmr{76.7}{35.2} & \cleanmr{73.0}{17.6} & \cleandelta{3.7}{17.6} & \cleanmr{77.3}{35.2} & \cleanmr{76.6}{18.3} & \cleandelta{0.7}{16.9} & \cleanmr{77.5}{47.2} & \cleanmr{73.2}{26.1} & \cleandelta{4.3}{21.1} & \cleandelta{2.9}{18.5} \\
\rowcolor{resultgroup}
\textbf{Average} & \cleanmr{75.1}{42.8} & \cleanmr{73.5}{23.9} & \cleandelta{1.6}{18.9} & \cleanmr{74.7}{35.6} & \cleanmr{74.6}{21.7} & \cleandelta{0.1}{14.0} & \cleanmr{75.4}{49.6} & \cleanmr{74.1}{30.3} & \cleandelta{1.3}{19.3} & \cleandelta{1.0}{17.4} \\
\midrule[0.4pt]
\rowcolor{resultgroup}
\multicolumn{11}{c}{\textbf{\textit{RL-based}}} \\
\modelname{searchr1}{Search-R1} & \cleanmr{77.6}{52.0} & \cleanmr{77.5}{45.3} & \cleandelta{0.1}{6.7} & \cleanmr{80.9}{53.4} & \cleanmr{82.5}{45.6} & \cleandelta{-1.6}{7.8} & \cleanmr{81.7}{53.4} & \cleanmr{82.5}{43.6} & \cleandelta{-0.8}{9.8} & \cleandelta{-0.8}{8.1} \\
\rowcolor{resultstripe}
\modelname{qwen}{ZeroSearch-base} & \cleanmr{75.9}{53.7} & \cleanmr{82.2}{39.8} & \cleandelta{-6.3}{13.9} & \cleanmr{79.9}{40.4} & \cleanmr{81.4}{37.8} & \cleandelta{-1.5}{2.6} & \cleanmr{78.7}{55.1} & \cleanmr{78.1}{43.1} & \cleandelta{0.6}{12.0} & \cleandelta{-2.4}{9.5} \\
\modelname{redsearcher}{REDSearcher} & \cleanmr{81.8}{44.6} & \cleanmr{76.3}{29.9} & \cleandelta{5.5}{14.7} & \cleanmr{81.6}{45.2} & \cleanmr{78.7}{40.0} & \cleandelta{2.9}{5.2} & \cleanmr{78.2}{52.2} & \cleanmr{79.0}{38.4} & \cleandelta{-0.8}{13.8} & \cleandelta{2.5}{11.2} \\
\rowcolor{resultstripe}
\modelname{hunyuan}{Search-P1} & \cleanmr{79.1}{50.8} & \cleanmr{79.2}{39.4} & \cleandelta{-0.1}{11.4} & \cleanmr{81.2}{43.6} & \cleanmr{81.0}{35.7} & \cleandelta{0.2}{7.9} & \cleanmr{80.1}{54.6} & \cleanmr{76.2}{41.1} & \cleandelta{3.9}{13.5} & \cleandelta{1.3}{10.9} \\
\rowcolor{resultgroup}
\textbf{Average} & \cleanmr{78.6}{50.3} & \cleanmr{78.8}{38.6} & \cleandelta{-0.2}{11.7} & \cleanmr{80.9}{45.7} & \cleanmr{80.9}{39.8} & \cleandelta{0.0}{5.9} & \cleanmr{79.7}{53.8} & \cleanmr{79.0}{41.6} & \cleandelta{0.7}{12.3} & \cleandelta{0.2}{9.9} \\
\bottomrule[1.5pt]
\end{tabularx}
\vspace{-6pt}
\end{table}

%% file: sections/rs_table_searchqa.tex
% Main results table (layout adapted from CheckerBench Table 2).
% Self-contained: logo macros below; icons in figures/model-icons and figures/harness-icons.
% Requires: graphicx, xcolor, colortbl, booktabs, tabularx, array.
\providecommand{\brandlabel}[2]{%
  \raisebox{-0.15em}{\includegraphics[height=1em,width=1em,keepaspectratio]{#1}}%
  \hspace{0.35em}#2}
\providecommand{\modelname}[2]{\brandlabel{figures/model-icons/model-#1.png}{#2}}
\providecommand{\harnessname}[2]{\brandlabel{figures/harness-icons/#1.png}{#2}}

% Inline changes follow the supplied reference: orange increases, teal decreases.
\begin{table}[!htbp]
\centering
\scriptsize
\definecolor{resultheader}{HTML}{DDE6EE}
\definecolor{resultstripe}{HTML}{F5F7F9}
\definecolor{resultgroup}{HTML}{EBF0F4}
\definecolor{resultup}{HTML}{D95638}
\definecolor{resultdown}{HTML}{008D9C}
\definecolor{datasetname}{HTML}{365F91}
\newcommand{\metricheader}[1]{%
  \begin{tabular}[c]{@{}c@{}}#1\end{tabular}}
% Reserve the same space for every value and subscript so decimals align.
\newcommand{\metricvalue}[2]{%
  \mbox{#1#2}}
% Column header centred over the value (not the value+subscript block).
\newcommand{\colhead}[1]{%
  \mbox{#1}}
\newcommand{\metricchange}[3]{%
  \raisebox{-0.35ex}{\fontsize{5}{5}\selectfont
    \textcolor{#1}{\ensuremath{#2}#3}}}
\newcommand{\deltaup}[1]{\metricchange{resultup}{\uparrow}{#1}}
\newcommand{\deltadown}[1]{\metricchange{resultdown}{\downarrow}{#1}}
\newcommand{\deltaequal}{\metricchange{gray}{=}{0.00}}
\caption{SearchQA. Beginning/Middle/End: Word Position within the target evidence text. Strict/Soft cells: bold full-set clean accuracy; $\downarrow$ MR on closed-book-correct items (\%). Each $\Delta$ cell: bold clean-accuracy contrast and $\downarrow$ MR contrast, both Strict minus Soft (pp); Mean averages positions. The arrow identifies the MR component, not an accuracy drop or a sign.}
\label{tab:rs-searchqa}
\setlength{\tabcolsep}{2pt}
\renewcommand{\arraystretch}{1.22}
\begin{tabularx}{\textwidth}{l*{10}{>{\centering\arraybackslash}X}}
\toprule[1.5pt]
\rowcolor{resultheader}
\brandlabel{figures/model-icons/dataset.png}{\textcolor{datasetname}{\textbf{\textit{SearchQA}}}} & \multicolumn{3}{c}{\textbf{Beginning}} & \multicolumn{3}{c}{\textbf{Middle}} & \multicolumn{3}{c}{\textbf{End}} & \colhead{\textbf{Mean}} \\
\arrayrulecolor{gray!60}\cmidrule(lr){2-4}\cmidrule(lr){5-7}\cmidrule(lr){8-10}\cmidrule(lr){11-11}\arrayrulecolor{black}
\rowcolor{resultheader}
\textbf{Model} & \colhead{\textbf{Strict}} & \colhead{\textbf{Soft}} & \colhead{$\boldsymbol{\Delta}_{\mathbf{instr}}$} & \colhead{\textbf{Strict}} & \colhead{\textbf{Soft}} & \colhead{$\boldsymbol{\Delta}_{\mathbf{instr}}$} & \colhead{\textbf{Strict}} & \colhead{\textbf{Soft}} & \colhead{$\boldsymbol{\Delta}_{\mathbf{instr}}$} & \colhead{$\overline{\boldsymbol{\Delta}}_{\mathbf{instr}}$} \\
\midrule[0.6pt]
\rowcolor{resultgroup}
\multicolumn{11}{c}{\textbf{\textit{API-based}}} \\
\modelname{openai}{GPT-5.6} & \cleanmr{90.4}{13.9} & \cleanmr{90.0}{4.6} & \cleandelta{0.4}{9.3} & \cleanmr{89.2}{9.9} & \cleanmr{87.4}{3.6} & \cleandelta{1.8}{6.3} & \cleanmr{91.0}{23.0} & \cleanmr{87.1}{14.5} & \cleandelta{3.9}{8.5} & \cleandelta{2.0}{8.0} \\
\rowcolor{resultstripe}
\modelname{claude}{Claude Opus 4.6} & \cleanmr{95.1}{14.6} & \cleanmr{90.2}{7.1} & \cleandelta{4.9}{7.5} & \cleanmr{92.4}{17.0} & \cleanmr{86.0}{6.6} & \cleandelta{6.4}{10.4} & \cleanmr{91.3}{16.7} & \cleanmr{89.3}{9.0} & \cleandelta{2.0}{7.7} & \cleandelta{4.4}{8.5} \\
\modelname{qwen}{Qwen3.8-Max} & \cleanmr{91.5}{14.8} & \cleanmr{91.6}{5.9} & \cleandelta{-0.1}{8.9} & \cleanmr{90.3}{17.4} & \cleanmr{91.1}{8.9} & \cleandelta{-0.8}{8.5} & \cleanmr{88.0}{24.3} & \cleanmr{85.9}{10.3} & \cleandelta{2.1}{14.0} & \cleandelta{0.4}{10.5} \\
\rowcolor{resultstripe}
\modelname{deepseek}{DeepSeek-V4-Pro} & \cleanmr{90.3}{18.1} & \cleanmr{88.2}{5.7} & \cleandelta{2.1}{12.4} & \cleanmr{87.7}{16.4} & \cleanmr{88.4}{5.9} & \cleandelta{-0.7}{10.5} & \cleanmr{90.1}{22.7} & \cleanmr{91.0}{9.7} & \cleandelta{-0.9}{13.0} & \cleandelta{0.2}{12.0} \\
\modelname{glm}{GLM-5.2} & \cleanmr{83.7}{18.1} & \cleanmr{86.3}{9.2} & \cleandelta{-2.6}{8.9} & \cleanmr{86.5}{16.5} & \cleanmr{87.3}{4.7} & \cleandelta{-0.8}{11.8} & \cleanmr{88.4}{26.6} & \cleanmr{88.2}{11.4} & \cleandelta{0.2}{15.2} & \cleandelta{-1.1}{12.0} \\
\rowcolor{resultstripe}
\modelname{kimi}{Kimi-K3} & \cleanmr{89.2}{15.9} & \cleanmr{89.1}{10.3} & \cleandelta{0.1}{5.6} & \cleanmr{83.3}{15.8} & \cleanmr{88.4}{8.1} & \cleandelta{-5.1}{7.7} & \cleanmr{88.4}{20.0} & \cleanmr{93.7}{9.0} & \cleandelta{-5.3}{11.0} & \cleandelta{-3.4}{8.1} \\
\modelname{hunyuan}{HY4-Preview} & \cleanmr{85.9}{25.3} & \cleanmr{86.8}{9.1} & \cleandelta{-0.9}{16.2} & \cleanmr{84.1}{22.7} & \cleanmr{84.4}{11.3} & \cleandelta{-0.3}{11.4} & \cleanmr{81.4}{26.6} & \cleanmr{87.6}{14.1} & \cleandelta{-6.2}{12.5} & \cleandelta{-2.5}{13.4} \\
\rowcolor{resultgroup}
\textbf{Average} & \cleanmr{89.4}{17.2} & \cleanmr{88.9}{7.4} & \cleandelta{0.6}{9.8} & \cleanmr{87.6}{16.5} & \cleanmr{87.6}{7.0} & \cleandelta{0.1}{9.5} & \cleanmr{88.4}{22.8} & \cleanmr{89.0}{11.1} & \cleandelta{-0.6}{11.7} & \cleandelta{0.0}{10.3} \\
\midrule[0.4pt]
\rowcolor{resultgroup}
\multicolumn{11}{c}{\textbf{\textit{Open-source}}} \\
\modelname{meta}{Llama-3-8B} & \cleanmr{75.8}{34.3} & \cleanmr{78.4}{23.9} & \cleandelta{-2.6}{10.4} & \cleanmr{71.4}{33.8} & \cleanmr{79.8}{21.1} & \cleandelta{-8.4}{12.7} & \cleanmr{73.4}{44.8} & \cleanmr{73.3}{31.7} & \cleandelta{0.1}{13.1} & \cleandelta{-3.6}{12.1} \\
\rowcolor{resultstripe}
\modelname{mistral}{Mistral-7B} & \cleanmr{80.9}{36.2} & \cleanmr{75.5}{22.7} & \cleandelta{5.4}{13.5} & \cleanmr{77.2}{37.6} & \cleanmr{79.4}{20.8} & \cleandelta{-2.2}{16.8} & \cleanmr{76.0}{48.7} & \cleanmr{76.5}{29.1} & \cleandelta{-0.5}{19.6} & \cleandelta{0.9}{16.6} \\
\modelname{qwen}{Qwen3-8B} & \cleanmr{80.1}{37.9} & \cleanmr{78.1}{21.0} & \cleandelta{2.0}{16.9} & \cleanmr{80.0}{27.6} & \cleanmr{81.0}{17.8} & \cleandelta{-1.0}{9.8} & \cleanmr{75.7}{38.4} & \cleanmr{79.7}{20.8} & \cleandelta{-4.0}{17.6} & \cleandelta{-1.0}{14.8} \\
\rowcolor{resultstripe}
\modelname{gemma}{Gemma-2-9B} & \cleanmr{77.0}{29.1} & \cleanmr{79.0}{21.1} & \cleandelta{-2.0}{8.0} & \cleanmr{73.3}{24.0} & \cleanmr{81.1}{18.5} & \cleandelta{-7.8}{5.5} & \cleanmr{78.7}{31.5} & \cleanmr{79.7}{22.7} & \cleandelta{-1.0}{8.8} & \cleandelta{-3.6}{7.4} \\
\rowcolor{resultgroup}
\textbf{Average} & \cleanmr{78.5}{34.4} & \cleanmr{77.8}{22.2} & \cleandelta{0.7}{12.2} & \cleanmr{75.5}{30.8} & \cleanmr{80.3}{19.6} & \cleandelta{-4.9}{11.2} & \cleanmr{76.0}{40.9} & \cleanmr{77.3}{26.1} & \cleandelta{-1.4}{14.8} & \cleandelta{-1.8}{12.7} \\
\midrule[0.4pt]
\rowcolor{resultgroup}
\multicolumn{11}{c}{\textbf{\textit{RL-based}}} \\
\modelname{searchr1}{Search-R1} & \cleanmr{80.8}{50.9} & \cleanmr{82.3}{37.6} & \cleandelta{-1.5}{13.3} & \cleanmr{84.2}{47.1} & \cleanmr{83.6}{33.1} & \cleandelta{0.6}{14.0} & \cleanmr{81.3}{51.7} & \cleanmr{84.1}{39.4} & \cleandelta{-2.8}{12.3} & \cleandelta{-1.2}{13.2} \\
\rowcolor{resultstripe}
\modelname{qwen}{ZeroSearch-base} & \cleanmr{78.1}{45.5} & \cleanmr{80.0}{37.9} & \cleandelta{-1.9}{7.6} & \cleanmr{80.9}{42.8} & \cleanmr{78.6}{30.3} & \cleandelta{2.3}{12.5} & \cleanmr{81.1}{53.2} & \cleanmr{82.8}{38.8} & \cleandelta{-1.7}{14.4} & \cleandelta{-0.4}{11.5} \\
\modelname{redsearcher}{REDSearcher} & \cleanmr{83.4}{38.8} & \cleanmr{86.5}{31.8} & \cleandelta{-3.1}{7.0} & \cleanmr{83.9}{32.0} & \cleanmr{83.3}{27.1} & \cleandelta{0.6}{4.9} & \cleanmr{82.2}{46.0} & \cleanmr{83.3}{36.7} & \cleandelta{-1.1}{9.3} & \cleandelta{-1.2}{7.1} \\
\rowcolor{resultstripe}
\modelname{hunyuan}{Search-P1} & \cleanmr{77.8}{45.6} & \cleanmr{85.2}{41.5} & \cleandelta{-7.4}{4.1} & \cleanmr{84.7}{42.4} & \cleanmr{80.3}{34.7} & \cleandelta{4.4}{7.7} & \cleanmr{83.2}{50.4} & \cleanmr{80.2}{36.7} & \cleandelta{3.0}{13.7} & \cleandelta{0.0}{8.5} \\
\rowcolor{resultgroup}
\textbf{Average} & \cleanmr{80.0}{45.2} & \cleanmr{83.5}{37.2} & \cleandelta{-3.5}{8.0} & \cleanmr{83.4}{41.1} & \cleanmr{81.5}{31.3} & \cleandelta{2.0}{9.8} & \cleanmr{82.0}{50.3} & \cleanmr{82.6}{37.9} & \cleandelta{-0.7}{12.4} & \cleandelta{-0.7}{10.1} \\
\bottomrule[1.5pt]
\end{tabularx}
\vspace{-6pt}
\end{table}

%% file: sections/robustness_protocol_appendix.tex
\section{Additional Position and Prompt Robustness Protocols}
\label{app:robustness-protocols}

This appendix accompanies Tables~\ref{tab:word-position-paired} and~\ref{tab:prompt-paraphrase}: 300 QA questions per model, with $|K_A|=220$ and $|K_B|=198$. QA, A, and B retain their supplied identifiers. These checks complement the main evaluation of fifteen systems and the 500-item audit; their contrasts and intervals use the corresponding QA records.

\subsection{Paired Word Position Analysis}
\label{sec:word-position-paired-protocol}

\paragraph{Required records and fixed population.}
Paired analysis requires frozen question IDs, references and aliases, foils,
closed-book responses, and all six clean/edited instruction--Word Position pairs.
The audit trail comprises model and chat-template versions, complete requests,
decoding settings, input lengths, and truncation records. Valid Word Position
comparisons hold document order fixed and verify the realized answer-span
locations; the supplementary audit's document slots are not a substitute.
$K_M$ remains fixed across conditions. Failed calls require recovery or explicit
accounting for missing data, not labels indicating incorrect answers. Discarding incomplete grids
or changing denominators by position changes the paired estimand.

\paragraph{Paired estimands.}
For a fixed model and dataset, let $h_{q,\pi,p}$ indicate an edited response that matches the designated foil but not the reference. The two position contrasts are
\[
G_{\pi}^{E-M}=\frac{100}{|K_M|}\sum_{q\in K_M}
 (h_{q,\pi,\mathrm{End}}-h_{q,\pi,\mathrm{Mid}}),\qquad
G_{\pi}^{B-M}=\frac{100}{|K_M|}\sum_{q\in K_M}
 (h_{q,\pi,\mathrm{Beg}}-h_{q,\pi,\mathrm{Mid}}).
\]
The third contrast is $G_{\pi}^{B-E}=G_{\pi}^{B-M}-G_{\pi}^{E-M}$, calculated within question before resampling; its interval is not obtained by subtracting interval endpoints. Table~\ref{tab:word-position-paired} reports all three. Under Strict, End-minus-Middle intervals are positive and Beginning-minus-End intervals are negative for both models. Beginning-minus-Middle remains unresolved, as do all Soft contrasts. These results for individual models do not establish the complete ordering or an instruction--position interaction.

\paragraph{Uncertainty and supplementary diagnostics.}
The supplied position results use 2,000 paired bootstrap resamples, retaining all repeated cells of a sampled question. The summary does not specify its random seed, method of constructing intervals, or multiplicity adjustment. Intervals are interpreted individually, not as simultaneous confirmation of every contrast. These analyses follow inspection of the main summaries and are not claimed as preregistered. A interval containing zero does not establish equality; a nonzero contrast does not identify an attention mechanism. Complete clean/edited records also permit direct transitions and comparisons of gold retention on the same $K_M$, but these require their own calculations. Clean accuracy on the full set must not be subtracted from MR, and the document-slot audit does not replicate this position intervention.

\subsection{Source Policy Paraphrases and Execution}
\label{sec:paraphrase-clauses}

\paragraph{Controlled prompt construction.}
Appendix~\ref{sec:prompt-templates} gives the original clauses specifying source policy. The candidates below paraphrase only that policy: Strict prioritizes documents despite apparent errors; Soft permits prior knowledge to override conflicting documents. Verify these meanings and freeze the clauses and cohort before inference. Insert one clause at the same location in the system prompt, holding the role, rules for selecting answers, output contract, wrapper for the user message, and evidence fixed. Add no verification, abstention, or explanation requests.

\begin{tcolorbox}[colback=green!5!white,colframe=green!60!blue,
title=Source-Policy Paraphrase Candidates,fonttitle=\bfseries,
boxrule=0.6pt,arc=1mm,left=2mm,right=2mm,top=1.5mm,bottom=1.5mm]
\small\raggedright
\textbf{Paraphrase 1: Strict}\par
Treat the supplied documents as the primary authority for your answer. Give a concise answer even when their contents seem incorrect.
\par\smallskip
\textbf{Paraphrase 1: Soft}\par
Consult the documents when useful. If their claims contradict well-known facts, give priority to your own knowledge.
\par\smallskip
\textbf{Paraphrase 2: Strict}\par
Base your concise answer primarily on the provided documents, including when the information in them appears to be mistaken.
\par\smallskip
\textbf{Paraphrase 2: Soft}\par
Draw on the supplied documents where they help, but rely on your own knowledge when they disagree with well-known facts.
\end{tcolorbox}

\paragraph{Execution and audit trail.}
The full design crosses three prompt versions, two policies, three Word Positions,
and two evidence arms: 36 requests per question and model. Reusing original
outputs is valid only with identical requests, model versions, and settings; the
two additional versions then require 24 new requests. The original closed-book
stratum remains fixed. Resource-limited subsets require selection before
inference. Reproduction requires variant IDs, complete messages, context IDs,
model revisions, generation settings, raw responses, matcher version, and
failure/retry records. The design includes neither live retrieval nor model updates.

\paragraph{Prompt contrasts and changes in effect size.}
Let $h^{(v)}_{q,\pi,p}$ denote the foil-only indicator under version $v$, with $v=0$ the original. For each question define
\[
\begin{aligned}
d_q^{(v)}&=\frac{1}{3}\sum_{p}
 \left(h^{(v)}_{q,\mathrm{Strict},p}-h^{(v)}_{q,\mathrm{Soft},p}\right),\\
\Delta_v&=\frac{100}{|K_M|}\sum_{q\in K_M}d_q^{(v)},\qquad
R_v=\frac{100}{|K_M|}\sum_{q\in K_M}(d_q^{(v)}-d_q^{(0)}).
\end{aligned}
\]
Comparisons pair questions across versions, policies, and positions. Table~\ref{tab:prompt-paraphrase} reports MR, $\Delta_v$, $R_v$, and 95\% intervals; its resample count and seed are unspecified (2,000 resamples applies only to the position table). The change from the original to itself is zero. All six instruction-contrast intervals exclude zero; all four changes from the original include zero. This supports consistent direction, not equivalence, which requires a prespecified meaningful margin and a dedicated analysis. Contrasts retain their supplied precision rather than being recomputed from rounded MR. Saved requests are needed to verify exact paraphrase bindings to execution messages.